\documentclass[journal]{IEEEtran}
\usepackage{amsmath,amsfonts}
\usepackage{algorithmic}
\usepackage{algorithm}
\usepackage{array}

\usepackage{cite}

\usepackage{textcomp}
\usepackage{stfloats}
\usepackage{url}
\usepackage{verbatim}
\usepackage{graphicx}
\usepackage{booktabs}
\usepackage{tikz}
\usepackage{dsfont}
\usepackage{bm}
\usepackage{subfigure}

\usepackage{tabularx}
\usepackage{pifont}
\usepackage{amsthm}

\begin{document}

\title{Efficient and Scalable Granular-ball Graph Coarsening Method for Large-scale Graph Node Classification}

\author{Guan~Wang,
        Shuyin~Xia$^{*}$,
        Lei~Qian,
        Tao~Wu,
        Guoyin~Wang,
        Yi~Wang,
        Wei~Wang
\thanks{Guan Wang, Shuyin Xia, and Lei Qian are with the School of Computer Science and Technology, Chongqing Key Laboratory of Computational Intelligence, Key Laboratory of Cyberspace Big Data Intelligent Security, Ministry of Education, Sichuan-Chongqing Co-construction Key Laboratory of Digital Economy Intelligence and Key Laboratory of Big Data Intelligent Computing, Chongqing University of Posts and Telecommunications, 400065, Chongqing, China. Tao Wu is with the School of Computer Science and Technology, School of Cyber Security and Information Law, Chongqing University of Posts and Telecommunications, 400065, Chongqing, China. Guoyin Wang is with Chongqing Key Laboratory of Brain-Inspired Cognitive Computing and Educational Rehabilitation for Children with Special Needs, Chongqing Normal University, Chongqing, 401331, China. E-mail: wangguan66@foxmail.com, xiasy@cqupt.edu.cn (corresponding author), 501\_5@163.com, wutao@cqupt.edu.cn, wanggy@cqnu.edu.cn.
Yi Wang and Wei Wang are with the Chongqing Ant Consumer Finance Co,. Ltd, Ant Group. E-mail: haonan.wy@myxiaojin.cn,wangshi.ww@myxiaojin.cn.}}




\markboth{Journal of \LaTeX\ Class Files,~Vol.~14, No.~8, August~2021}%
{Shell \MakeLowercase{\textit{et al.}}: A Sample Article Using IEEEtran.cls for IEEE Journals}


\maketitle

\begin{abstract}
Graph Convolutional Network (GCN) is a model that can effectively handle graph data tasks and has been successfully applied. However, for large-scale graph datasets, GCN still faces the challenge of high computational overhead, especially when the number of convolutional layers in the graph is large. Currently, there are many advanced methods that use various sampling techniques or graph coarsening techniques to alleviate the inconvenience caused during training. However, among these methods, some ignore the multi-granularity information in the graph structure, and the time complexity of some coarsening methods is still relatively high. In response to these issues, based on our previous work, in this paper, we propose a new framework called Efficient and Scalable Granular-ball Graph Coarsening Method for Large-scale Graph Node Classification. Specifically, this method first uses a multi-granularity granular-ball graph coarsening algorithm to coarsen the original graph to obtain many subgraphs. The time complexity of this stage is linear and much lower than that of the exiting graph coarsening methods. Then, subgraphs composed of these granular-balls are randomly sampled to form minibatches for training GCN. Our algorithm can adaptively and significantly reduce the scale of the original graph, thereby enhancing the training efficiency and scalability of GCN. Ultimately, the experimental results of node classification on multiple datasets demonstrate that the method proposed in this paper exhibits superior performance. The code is available at https://anonymous.4open.science/r/1-141D/.
\end{abstract}


\section{Introduction}
\IEEEPARstart{G}{raph} is a network based on entities and their relationships, capable of describing more complex relationships among entities. In the real world, many complex systems can be described and analyzed using graphs, such as social networks, literature citation networks, protein interaction networks, urban transportation networks, etc. In recent years, graph neural networks (GNNs) \cite{fan2023generalizing, zhang2018arbitrary, besta2024parallel, guo2024gnn, ju2025survey} have demonstrated great potential in various fields, including social network analysis \cite{zhang2022improving, iyer2024non, zhou2025lgb}, drug discovery \cite{shiokawa2024efficient}, recommendation systems \cite{gao2022graph, shi2025noise, bao2024improved, liang2024survey}, etc., greatly promoting tasks such as classification and clustering \cite{guo2017improved, wu2020comprehensive, liang2024survey}. The core mechanism of GNN models is message passing, which enables nodes in the graph to aggregate and update information with neighboring nodes.

Graph convolutional networks (GCNs) \cite{bruna2013spectral, kipf2016semi, wu2019simplifying, bo2020structural} are neural network models that migrate from the image domain to the graph domain, and they can effectively handle graph-structured data. Its work has attracted extensive attention from researchers and has demonstrated outstanding performance in many application fields such as semi-supervised node classification \cite{kipf2016semi}, recommendation systems \cite{xu2025cohesion}, and traffic prediction \cite{zhang2024irregular}. GCN mainly gradually extracts the advanced features of nodes and learns the representations of nodes through multi-layer graph convolution operations. The graph convolution operation at each layer updates the representation of the node through the neighbor information of the node. For large-scale graph datasets, on the one hand, the traditional approach of performing graph convolution across the entire graph is often unrealistic. On the other hand, as the number of layers in the GCN network deepens, the number of neighbors that need to be traced back increases, and thus the training time will also grow exponentially. This poses challenges to the training of GCN of large-scale graph datasets.

\begin{figure}[!t]
\centering
\includegraphics[width=3.0in]{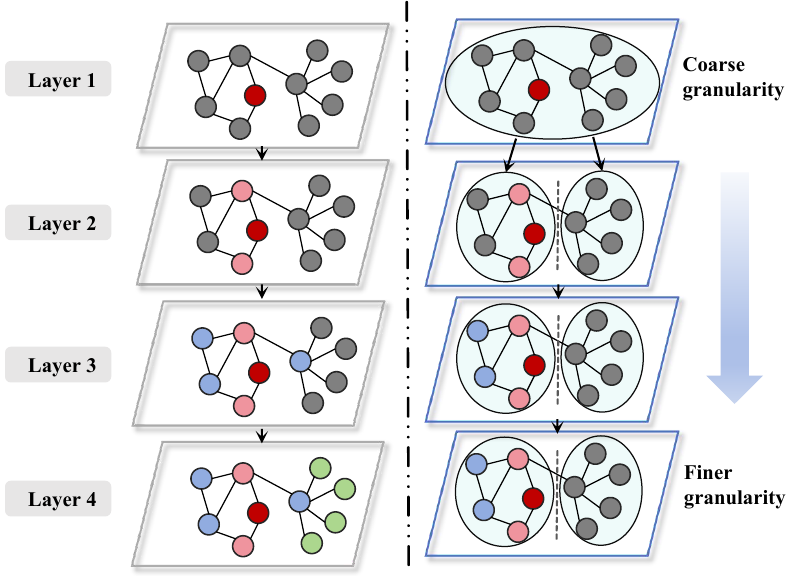}
\caption{The neighborhood expansion difference between traditional GCN and our proposed method. The red node is the starting node for neighborhood nodes expansion. The traditional GCN has an exponential neighborhood expansion problem, while our method can limit the scope of neighborhood expansion by capturing multi-granularity structural information in the graph from coarse-grained to fine-grained.}
\label{motivationfig}
\end{figure}

To avoid the above problems, there are some methods based on sampling technology \cite{hamilton2017inductive, chen2018stochastic, chen2018fastgcn, huang2018adaptive, chiang2019cluster, zenggraphsaint}, which use sampling and aggregation strategies based on GCN. Another category of methods involves reducing the scale of graphs, primarily through two approaches: graph condensation \cite{jin2021graph, huang2021scaling, pang2021graph}, which aims to learn a smaller synthetic graph that can replace the original graph as much as possible in downstream tasks. The other method is graph coarsening \cite{xia2025graph}, which differs from graph condensation. This approach aims to reduce the size of the graph by merging nodes with similar structures into supernodes, while preserving as much as possible the structural properties of the original graph.

Sampling-based techniques primarily focus on proximity nodes characterized by a single granularity of neighborhood relations with fixed hop counts and fixed quantities, where the fixed hop count somewhat limits their ability to capture deep-level structures. Additionally, the fixed subgraph structure formed by fixed hop counts and fixed node numbers overlooks the inherent multi-granularity characteristics of the graph structure (as shown in Figure \ref{motivationfig}) and limits its ability to capture complex relationships between nodes. The training method based on graph condensation may be affected by network performance and is only effective under a fixed condensation control rate. In recent years, some new multi-granularity graph coarsening methods, called granular-ball graph coarsening, such as SGBGC \cite{xia2025graph} and GBGC \cite{GBGC}, have divided subgraphs whose size is not limited by a fixed number of nodes or hops. Therefore, it can better capture the inherent multi-granularity characteristics of the graphic structure, and the entire coarsening process has good adaptability, requiring only the setting of one hyperparameter, namely the purity threshold. Although the granular-ball graph coarsening method has achieved good efficiency with a time complexity of $O(N^\frac{3}{2}+E\sqrt{N})$, it is still not low enough to be extended to large-scale graph dataset processing problems. 

To this end, this paper presents an extended framework called Efficient and Scalable Granular-ball Graph Coarsening Method for Large-scale Graph Node Classification (GB-CGNN). More specifically, this method innovatively introduces the METIS algorithm \cite{karypis1998fast} to generate initial granular-balls in the coarse-grained initialization stage. In the fine-grained binary splitting stage, an adaptive splitting mechanism is still used to refine these initial granular-balls into high-quality supernodes. This adaptability enhances the flexibility of graph coarsening. The new method GB-CGNN proposed in this paper reduces the time complexity of the granular-ball graph coarsening algorithm to a linear level for the first time, which is significantly lower than GBGC \cite{GBGC} and SGBGC \cite{xia2025graph}. 

In conclusion, the contributions of this paper are summarized as follows:

\begin{itemize}
  \item For the first time, we have reduced the time complexity of the granular-ball graph coarsening method to linear ($O(N)$), and the coarsening time is much lower than that of the existing state-of-the-art (SOTA) methods.
  \item By using the multi-granularity subgraphs corresponding to granular-ball supernodes as input for GCN, instead of sampling fine-grained point-level or single-granularity neighborhood relations, the training process is significantly accelerated. Additionally, the multi-granularity neighborhood relation sampling enables GB-CGNN to achieve better learning accuracy in many scenarios.
  \item The effectiveness and accuracy of GB-CGNN have been confirmed through large-scale testing and dataset comparison verification. This provides a new perspective for both graph coarsening and GCN training methods.
\end{itemize}

\section{Related work}
\subsection{Sampling methods for GCN training}
Bruna et al. \cite{bruna2013spectral} first extended the convolution idea to the field of graphs. Kipf et al. \cite{kipf2016semi} encourage the selection of convolutional architectures through local first-order approximations of spectral convolution. They mainly target relatively small datasets, thus enabling full-batch training. Later, in order to extend the GCN to large graphs, several classic inductive network embedding methods based on the GCN framework were proposed, and it was proved that the sampling technique is useful in small-batch training. Examples include GraphSAGE \cite{hamilton2017inductive}, FastGCN \cite{chen2018fastgcn}, VR-GCN \cite{chen2018stochastic} and AS-GCN \cite{huang2018adaptive}, etc. These methods sample the adjacent nodes of the target node and apply different aggregation strategies to obtain the feature vectors of these adjacent nodes, ultimately achieving the purpose of generating the embedding of the target node. Specifically, GraphSAGE performs uniform node sampling on the neighbors of the previous layer. It pre-defines the sample size budget in advance, thereby limiting the complexity of small-batch computations and enabling distributed training. In general, the main idea of this method is to learn a function that generates the embedding of a node by sampling and aggregating features from the local neighborhood of the node. Ying et al. \cite{ying2018graph} improved the sampling strategy of GraphSAGE by introducing an importance score for each neighbor. FastGCN \cite{chen2018fastgcn} accelerates the GCN training process by using importance sampling technology. Although importance sampling can reduce variance, small batches may become too sparse and fail to achieve the expected accuracy. AS-GCN \cite{huang2018adaptive} improves FastGCN through an additional sampling neural network. This method samples based on the selected nodes in the lower layer to achieve high accuracy. However, because of the expensive sampling algorithm and the additional parameter of the sampler to be learned, the computational cost may increase. VR-GCN \cite{chen2018stochastic} further limits the size of the neighborhood and suggests setting the number of neighbors sampled to 2. 

Some other researchers' work did not sample the layers but instead constructed minibatches through subgraphs. For instance, Chiang et al. \cite{chiang2019cluster} proposed minibatch training based on graph clustering. This method first performs clustering operations on the training graph to obtain a specified number of clusters, and then randomly selects clusters as minibatches for training, accelerating the GCN training process. Zeng et al. \cite{zenggraphsaint} introduced a graph neural network model based on sampling graphs. The computations on each minibatch are carried out on the sampling graphs, avoiding the ``neighbor explosion'' phenomenon and theoretically ensuring unbiasedness and minimum variance. 

\subsection{Graph coarsening and graph condensation methods for GCN training}
Because the training process of GNNs is usually costly, methods such as graph coarsening and graph condensation have received increasing attention. These methods aim to train GNNs through smaller coarsened graphs and synthetic graphs, thereby reducing costs to a certain extent. Among them, the SCAL algorithm \cite{huang2021scaling} demonstrates the scalable training of GNN through graph coarsening technology, proving that even if the graph nodes are reduced to one-tenth of the original size, there is no significant impact on the classification accuracy. Kumar et al. \cite{kumar2023featured} designed an optimity-based graph coarsening framework FGC, which takes the adjacency matrix and node features as input and jointly learns the coarsened versions of the graph and feature matrix to ensure the desired attributes while optimizing the graph representation. Jin et al. \cite{jin2021graph, jin2022condensing} proposed the graph condensation method GCOND, which not only reduces the number of nodes but also compresses the node features and structural information in a supervised manner. CMGC \cite{dickens2024graph} is also a training-dependent coarsening method that focuses on preserving graph convolution operations. The SGBGC proposed by Xia et al. \cite{xia2025graph} significantly enhances the scalability of networks such as GNN and GCN by coarsening the original graph by more than ten times without losing accuracy.

\subsection{Granular-ball computing}
Chen pointed out that human cognition exhibits the law of ``global precedence'' in his research published in Sciences \cite{chen1982topological}. Inspired by the law, Xia and Wang et al. introduced an innovative computational method known as granular-ball computing, celebrated for its efficiency and robustness \cite{xia2023granular}. The fundamental concept of this method involves initially treating the entire dataset as a granular-ball, which is subsequently split into smaller granular-balls. In granular-ball computing, granular-balls of varying sizes are employed to represent the sample space, upon which subsequent calculations are based. In contrast to traditional methods that utilize the most fine-grained points as input, granular-ball computing operates with granular-balls as input, rendering it efficient, robust, and interpretable \cite{xia2023granular}. Some existing granular-ball computing methods are mainly used in discrete data. Given a dataset $\mathcal{D}$, $\mathcal{GB}_j(j=1,2,\cdots,K)$ denotes the granular-ball generated based on $\mathcal{D}$, $K$ is the number of granular-balls. $\mathcal{GB}_j=\{g_i|i=1,2,\cdots,m_j\}$, $m_j$ denotes the number of data points contained in the $\mathcal{GB}_j$. The center of $\mathcal{GB}_j$ is defined as $c_j=\frac{1}{m_j}\sum_{i=1}^{m_j}g_i$, radius is defined as $r_j=\frac{1}{m_j}\sum_{i=1}^{m_j}(\lvert g_i-c_j\rvert)$. When the qualities of these granular-balls fail to meet the given quality threshold $T$, it becomes imperative to further split them into smaller granular-balls until the quality of each granular-ball meets the $T$. This iterative process ensures algorithm convergence, resulting in the clearest decision boundary. Recently, granular-ball computing has been applied across various domains of artificial intelligence, including rough sets, clustering, reinforcement learning, and graph neural networks, as demonstrated by recent works \cite{xia2020ball, xia2023gbrs, xia2024gbsvm, liu2024unlock, xie2025aw, GBGC, xia2025graph, Quantum, KernelK-means, xia2025adaptive}.

\section{Preliminaries}
\subsection{Graph coarsening}

\begin{table*}[tb!]
  \caption{Table of Notations.}
  \label{Notation}
  \begin{center}
  \begin{tabular}{cc|cc}
    \toprule
    \textbf{Notation} & \textbf{Meaning} & \textbf{Notation} & \textbf{Meaning} \\
    \midrule
    $\mathcal{D}$ & A dataset composed of discrete data points & $\mathcal{\overline{G}}$ & The coarsened graph of $\mathcal{G}$ \\
    $\mathcal{GB}$ & Granular-ball & $\mathcal{\overline{V}}$ & The node set of $\mathcal{\overline{G}}$ \\
    $K$ & The number of $\mathcal{GB}$ generated based on $\mathcal{D}$ & $\mathcal{\overline{E}}$ & The edge set of $\mathcal{\overline{G}}$ \\
    $g$ & The discrete data points contained in $\mathcal{GB}$ & $R$ & Rayleigh quotient \\
    $m$ & The number of discrete data points contained in $\mathcal{GB}$ & $\overline{N}$ & The number of nodes in $\mathcal{\overline{G}}$ \\
    $c$ & The center of $\mathcal{GB}$ & $\overline{E}$ & The number of edges in $\mathcal{\overline{G}}$ \\
    $r$ & The radius of $\mathcal{GB}$ & $\overline{v}$ & The node in $\mathcal{\overline{G}}$ \\
    $T$ & The quality threshold & $\mathbf{\overline{A}}$ & The adjacency matrix of $\mathcal{\overline{G}}$ \\
    $\mathcal{G}$ & The original graph & $\mathbf{\overline{C}}$ & The projection matrix that maps nodes in $\mathcal{V}$ to supernodes in $\mathcal{\overline{V}}$ \\
    $\mathcal{V}$ & The node set of $\mathcal{G}$ & $\mathbf{\overline{L}}$ & The Laplacian matrix of $\mathcal{\overline{G}}$ \\
    $\mathcal{E}$ & The edge set of $\mathcal{G}$ & $\mathcal{\widetilde{G}}$ & The subgraph ($\mathcal{GB}$) of $\mathcal{G}$ \\
    $N$ & The number of nodes in $\mathcal{G}$ & $\mathcal{\widetilde{V}}$ & The node set of $\mathcal{\widetilde{G}}$ \\
    $E$ & The number of edges in $\mathcal{G}$ & $\mathcal{\widetilde{E}}$ & The edge set of $\mathcal{\widetilde{G}}$ \\
    $v$ & The node in $\mathcal{G}$ & $\widetilde{N}$ & The number of nodes in $\mathcal{\widetilde{G}}$ \\
    $\mathbf{A}$ & The adjacency matrix of $\mathcal{G}$ & $\widetilde{E}$ & The number of edges in $\mathcal{\widetilde{G}}$ \\
    $\mathbf{D}$ & The degree matrix of $\mathcal{G}$ & $\mathbf{\widetilde{A}}$ & The adjacency matrix of $\mathcal{\widetilde{G}}$ \\
    $\mathbf{L}$ & The Laplacian matrix of $\mathcal{G}$ & $t$ & The number of subgraphs in $\mathcal{G}$ also indicates the number of $\mathcal{GB}$ in $\mathcal{\overline{G}}$ \\
    $\hat{\mathbf{L}}$ & The normalized Laplacian matrix of $\mathcal{G}$ & $L$ & The number of graph convolution layers \\
    $\mathbf{A}'$ & The normalized and regularized adjacency matrix of $\mathcal{G}$ & $\mathbf{X}$ & The feature matrix of $\mathcal{G}$ \\
    $\mathbf{W}$ & The weight matrix which will be learnt & $\mathbf{H}$ & The node embedding matrix \\
    $F$ & The dimension of node features & $y_i$ & The label of node $v_i$ in $\mathcal{G}$ \\
    $\hat{y}_i$ & The predicted label of node $v_i$ in $\mathcal{G}$
    & $N_t$ & The total number of test nodes \\
    $C$ & The number of classes & $k$ & The number of initial granular-balls \\
    $S$ & The number of cut edges in the METIS algorithm & $\mathbf{Y}$ & The label of nodes in $\mathcal{G}$ \\
    $n_i$ & The number of nodes with label $i$ in $\mathcal{GB}$ & & \\
    \bottomrule
  \end{tabular}
  \end{center}
\end{table*}


The notations and their corresponding meanings used
throughout the paper are summarized in Table \ref{Notation}. We denote an original graph as $\mathcal{G=(V,E)}$, whose node set is $\mathcal{V}$ with size $\lvert \mathcal{V} \rvert=N$, and whose edge set is $\mathcal{E}$ with size $\lvert \mathcal{E} \rvert=E$. The adjacency matrix of $\mathcal{G}$ is denoted as $\mathbf{A} \in \{{0,1}\}^{N \times N}$, and $\mathbf{A}_{ij}=1$ if there is an edge between nodes $v_i$ and $v_j$. The Laplacian matrix is denoted as $\mathbf{L}=\mathbf{D}-\mathbf{A}$, where $\mathbf{D}$ is the degree matrix with $\mathbf{D}_{ii}=\sum_j\mathbf{A}_{ij}$. The normalized Laplacian matrix is $\hat{\mathbf{L}}=\mathbf{D}^{-\frac{1}{2}}\mathbf{L}\mathbf{D}^{-\frac{1}{2}}$. Given an original graph $\mathcal{G}$, graph coarsening methods aim to find a smaller coarsened graph denoted as $\mathcal{\overline{G}=(\mathcal{\overline{V}},\mathcal{\overline{E}}})$ to approximate $\mathcal{G}$. The number of nodes in $\mathcal{\overline{G}}$ is denoted as $\overline{N}$, the number of edges is denoted as $\overline{E}$, and $\overline{N} < N$, $\overline{E} < E$. The adjacency matrix of $\overline{\mathcal{G}}$ is denoted as $\mathbf{\overline{A}} \in \{{0,1}\}^{\overline{N} \times \overline{N}}$, and $\mathbf{\overline{A}}_{ij}=1$ if there is an edge between supernodes $\overline{v}_i$ and $\overline{v}_j$. In the matrix form, we use $\mathbf{C} \in \mathbb{R}^{N \times \overline{N}}$ to denote a projection matrix that maps nodes in $\mathcal{V}$ to supernodes in $\mathcal{\overline{V}}$, and $\mathbf{C}_{ij}=1$ if node $i$ in $\mathcal{V}$ is mapped to supernode $j$ in $\mathcal{\overline{V}}$. The Laplacian matrix of the coarsened graph can be denoted as $\mathbf{\overline{L}}=\mathbf{C}^T\mathbf{L}\mathbf{C}$. This computation ensures that the spectral properties of the original graph are preserved in the coarsened graph. 

\subsection{Problem description}
Firstly, our proposed GB-CGNN seeks to obtain a coarsened graph $\mathcal{\overline{G}}$ to approximate $\mathcal{G}$ by using granular-ball graph coarsening. In other words, the original graph $\mathcal{G}$ is refined into multiple non-overlapping subgraphs (granular-balls) as follows:

\begin{align}
\mathcal{G}=\{\mathcal{\widetilde{G}}_1,\mathcal{\widetilde{G}}_2,\cdots,\mathcal{\widetilde{G}}_t\},t=\overline{N}, \\\mathcal{\widetilde{G}}_i=(\mathcal{\widetilde{V}}_i,\mathcal{\widetilde{E}}_i),i=1,2,\cdots,t,
\end{align}

\noindent where $\mathcal{\widetilde{V}}_i$ and $\mathcal{\widetilde{E}}_i$ represent the set of nodes and edges of the graph $\mathcal{\widetilde{G}}_i$ contained inside the granular-ball $\mathcal{GB}_i$, with $\mathcal{\widetilde{V}}_i\subset\mathcal{V}$ being a subset of nodes from $\mathcal{G}$. $\overline{N}$ is the number of supernodes in $\mathcal{\overline{G}}$, and each $\mathcal{GB}$ corresponds to a supernode in $\mathcal{\overline{G}}$. In the process of granular-ball graph coarsening implementation, the information loss between the original graph and the coarsened graph should be minimized, the number of granular-balls should be as few as possible, and the adaptive quality requirements should be met. 

Secondly, an $L$-layer GCN consists of $L$ graph convolution layers and each of them constructs embeddings for each node by mixing the embeddings of
neighbors of the node in the graph from the previous layer, mathematically:

\begin{equation}
\mathbf{H}^{(L+1)}=\mathbf{A}'\mathbf{X}^{(L)}\mathbf{W}^{(L)}, \mathbf{X}^{(L+1)}=\sigma(\mathbf{H}^{(L+1)}),
\label{h}
\end{equation}

\noindent where $\mathbf{A}'$ is the normalized and regularized adjacency matrix, $\mathbf{X}^{(l)}\in \mathbb{R}^{N\times F_l}$ is the embedding at the $l$-th layer for all $N$ nodes and $\mathbf{X}^{(0)}=\mathbf{X}$, $\mathbf{X} \in \mathbb{R}^{N\times F}$ is the feature matrix for all $N$ nodes, $\mathbf{W}^{(l)}$ is the weight matrix which will be learnt for the downstream tasks, and $\sigma(\cdot)$ is the $ReLU$ activation function \cite{nair2010rectified}. For
simplicity we assume the feature dimensions are the same for all layers, that is to say, $F_1=F_2=\cdots=F_L=F$. 

Finally, in the node classification task, each node $v_i$ in the original graph has a label $y_i$ ($y_i \in {0,1,2,\cdots,C-1}$), where $C$ is the number of classes. The classifier is initially aware of the labels of a subset of nodes $\mathcal{V_L}$. The goal of node classification is to infer the labels of the nodes in $\mathcal{V \setminus V_L}$ by learning a classification function from the coarsened graph $\mathcal{\overline{G}}$, effectively using the reduced graph structure to predict labels with reduced computational resources while maintaining or
even improving the accuracy of the prediction. When using GCN for this application, the goal is to learn
weight matrices in Formula (\ref{h}) by minimizing the following cross-entropy loss function:

\begin{equation}
\mathcal{L} = -\frac{1}{\lvert \mathcal{V_L} \rvert} \sum_{i \in \mathcal{V_L}} \sum_{c=1}^{C} y_{i,c}log(\hat{y}_{i,c}),
\end{equation}

\noindent where $\hat{y}_{i,c}$ denotes the predicted probability that node $v_i$ belongs to class $c$, $y_{i,c}$ is the one-hot representation of the real label of node $v_i$. 

\section{Methodology}

\begin{figure*}
  \includegraphics[width=\textwidth]{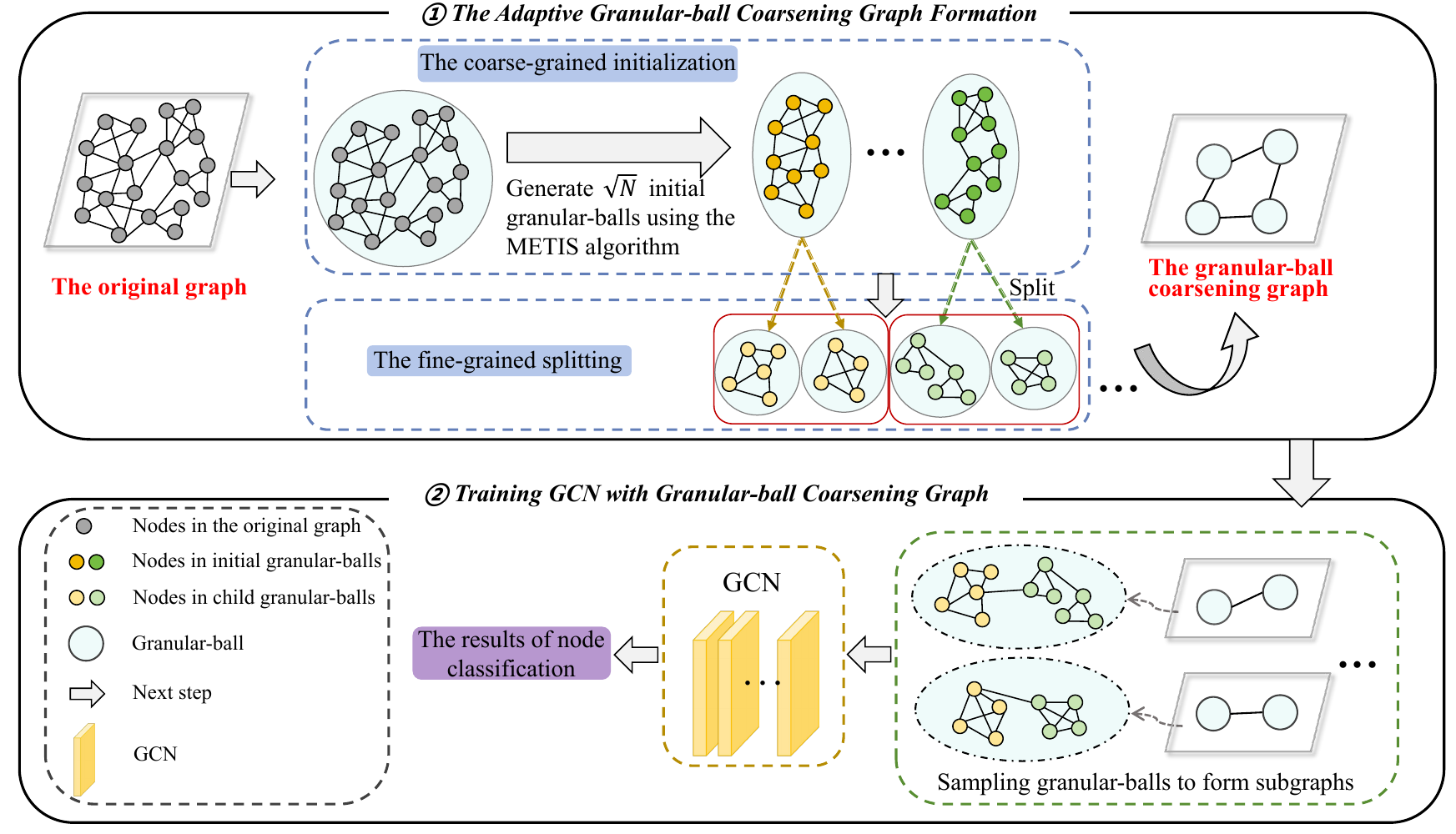}
  \caption{The framework of our proposed GB-CGNN. (1) In the \textit{Adaptive Granular-ball Coarsening Graph Formation module}, we split the coarsest granular-ball from coarse-grained to fine-grained based on the adaptive splitting condition to obtain the granular-ball coarsening graph, where each granular-ball corresponds to each subgraph of the original graph. (2) In the \textit{Training GCN with Granular-ball Coarsening Graph module}, we randomly sampled some granular-balls to form subgraphs and trained them with GCN, ultimately obtaining the results of node classification.}
  \label{framework}
\end{figure*}

In this section, we will present a detailed introduction to our proposed framework called Efficient and Scalable Granular-ball Graph Coarsening Method for Large-scale Graph Node Classification (GB-CGNN). Figure \ref{framework} illustrates the overall framework of GB-CGNN, which primarily comprises \textbf{two modules}: \textit{\textbf{(1)} The Adaptive Granular-ball Coarsening Graph Formation}. \textit{\textbf{(2)} Training GCN with Granular-ball Coarsening Graph}. Next, we introduce each module in detail.

\subsection{The Adaptive Granular-ball Coarsening Graph Formation}
This stage mainly consists of five steps. \textbf{Step 1}. Based on the law of ``global precedence'' \cite{chen1982topological}, the original graph as the coarsest granular-ball. \textbf{Step 2}. Generate $\sqrt{N}$ initial granular-balls using the METIS algorithm \cite{karypis1998fast}. The value of $\sqrt{N}$ is an empirical choice, referenced from previous work \cite{xie2020new, yu2001upper, xia2025graph, GBGC}, which has been shown to balance the trade-off between size and quantity of granular-balls. 

The METIS algorithm \cite{karypis1998fast} is an efficient algorithm for graph partitioning, typically applied in the division of large-scale graphs, especially in applications such as parallel computing, load balancing, and meshing. The core idea of METIS is to divide the graph into several subgraphs, making the nodes within each subgraph closely connected while reducing the number of edges between the subgraphs. The procedure for generating $\mathcal{GB}_{init}$ using the METIS algorithm is as follows. $\mathcal{GB}_{init}$ is described mathematically as follows:

\begin{equation}
\mathcal{GB}_{init}=\{\mathcal{GB}_1,\mathcal{GB}_2,\cdots,\mathcal{GB}_k\},k=\sqrt{N}. 
\end{equation}

When using METIS algorithm to divide the original graph $\mathcal{G}$ into $k$ ($k=\sqrt{N}$) subgraphs $\mathcal{G}_1,\mathcal{G}_2,\cdots,\mathcal{G}_k$, each subgraph $\mathcal{G}_i(i=1,2,\cdots,k)$ contains a portion of nodes $\mathcal{V}_i \subset \mathcal{V}$ and satisfies:

\begin{equation}
\bigcup_i\mathcal{V}_i=\mathcal{V}, \mathcal{V}_i \cap \mathcal{V}_j=\emptyset,\text{if} \; i\neq j.
\end{equation}

The number of edges between each subgraph should be as few as possible. Specifically, the optimization objective of METIS is to maintain load balancing while simultaneously minimizing the number of cut edges $S$. Mathematically:

\begin{equation}
\min_{\{\mathcal{V}_1,\mathcal{V}_2,\cdots,\mathcal{V}_k\}}\left(\sum_{i\neq j}|\mathcal{E}(\mathcal{V}_i,\mathcal{V}_j)|\right),
\end{equation}

\noindent where $\lvert \mathcal{E}(\mathcal{V}_i,\mathcal{V}_j)\rvert$ denotes the number of edges between subgraph $\mathcal{G}_i$ and subgraph $\mathcal{G}_j$. In addition to minimizing the number of cutting edges, METIS also needs to consider the issue of load balancing. That is to say, the size of each subgraph should be as equal as possible. Typically, the size of a subgraph is constrained by the weights of the nodes. 

On the one hand, for weighted graphs, if each node $v_i \in \mathcal{V}$ has a weight $w(v)$, the goal of METIS is to make the sum of the weights of each subgraph as equal as possible. Mathematically:

\begin{equation}
\lvert \sum_{v \in \mathcal{V}_i} w(v)-\frac{1}{k} \sum_{v \in \mathcal{V}}w(v) \rvert \leq \epsilon  \cdot \sum_{v \in \mathcal{V}}w(v),
\end{equation}

\noindent where $\epsilon$ is a small tolerance value, indicating that the load of each subgraph can have some deviation.

On the other hand, for unweighted graphs, ensuring load balancing mainly depends on the number of nodes rather than their weights. That is, the number of nodes in each subgraph should be as equal as possible. Mathematically:

\begin{equation}
\lvert \mathcal{V}_i \rvert - \frac{N}{k} \leq \sigma,
\end{equation}

\noindent where $\lvert \mathcal{V}_i \rvert$ is the number of nodes in subgraph $\mathcal{G}_i$, $\frac{N}{k}$ is the number of nodes for each subgraph in an ideal situation, and $\sigma$ is a pre-set constant, indicating the allowable deviation of the number of nodes. 

In general, the METIS algorithm leverages the spectral method based on the Laplacian matrix in conjunction with multi-level graph partitioning to minimize the number of cut edges while simultaneously maintaining load balancing, thereby facilitating efficient and effective graph partitioning.

All nodes participate in the calculation, so global calculation is involved in this step. The purpose of coarse-grained initial partitioning is twofold. Firstly, it leverages the global characteristics of the graph structure to ensure a more uniform and straightforward graph distribution within each initial granular-ball in $\mathcal{GB}_{init}$. Secondly, it provides a solid starting point for the subsequent fine-grained binary splitting process, enabling the granular-ball quality measurement to function more effectively.

\textbf{Step 3}. Calculate the quality of each granular-ball, which is defined as follows.

\textbf{Definition 1} For a granular-ball $\mathcal{GB}$, its quality is formulated as follows: 

\begin{equation}
quality(\mathcal{GB})=\frac{\widetilde{E}}{\widetilde{N}},
\end{equation}

\noindent where $\widetilde{E}$ and $\widetilde{N}$ represent the number of edges and nodes within the graph contained inside $\mathcal{GB}$, respectively.

Quality is defined of the average degree ($AD$), which quantifies the average number of edges connected to each node in the graph, serving as an intuitive measure of the connectivity within the graph. The higher the quality of the granular-ball, the more tightly connected the nodes inside the granular-ball. This method of quantifying the quality of the granular-ball ensures that the structural properties of the graph are considered and balanced in subsequent splitting of the granular-ball. 

\textbf{Step 4}. Perform a fine-grained binary splitting for each granular-ball $\mathcal{GB}_j(j=1,2,\cdots,k)$ in $\mathcal{GB}_{init}$. The fine-grained splitting identifies the two nodes with the highest degree of parent granular-ball ($\mathcal{GB}_j$) as the splitting centers of the two child granular-balls ($\mathcal{GB}_{j_{A}}$ and $\mathcal{GB}_{j_{B}}$), which are denoted as 
\begin{equation}
v_A=\arg\max\limits_{v\in\mathcal{\widetilde{V}}_{j}}deg(v), v_B=\arg\max\limits_{v\in\mathcal{\widetilde{V}}_{j}\setminus{v_A}}deg(v), 
\end{equation}

\noindent where $\mathcal{\widetilde{V}}_{j}$ represents the set of nodes within the parent granular-ball $\mathcal{GB}_j$. Based on the shortest path principle, each center node traverses the remaining nodes layer by layer through breadth first search (BFS), and other non-center nodes are assigned to the node according to who was first traversed by a center node’s BFS:

\begin{align}
v &\in \mathcal{GB}_{j_{A}} \Longleftrightarrow d(v,v_{A}) \leq d(v,v_{B}), \\
v &\in \mathcal{GB}_{j_{B}} \Longleftrightarrow d(v,v_{B}) < d(v,v_{A}).
\end{align}

Only the nodes within granular-balls are involved in the calculation, so Steps 3 and 4 involve local calculation. The adaptive splitting condition of parent granular-ball splitting is given by the following formula:

\begin{equation}
({quality}(\mathcal{GB}_{j_{A}}+\mathcal{GB}_{j_{B}}))/2 > {quality}(\mathcal{GB}_j),
\end{equation}
\noindent where ${quality}(\mathcal{GB}_{j_{A}}+\mathcal{GB}_{j_{B}})$ denotes the sum of the qualities of the two child granular-balls, and $quality(\mathcal{GB}_j)$ denotes the quality of the parent granular-ball. This formula is employed to decide whether to proceed with splitting granular-balls. If there is a significant difference between the average quality of child granular-balls and that of the parent granular-ball, suggesting that further splitting improves the overall system performance, the splitting operation will continue. This strategy ensures both the precision and efficiency of the splitting process, preventing unnecessary complexity from arising.

\textbf{Step 5}. Generate results of granular-ball coarsening graph. In the process of graph coarsening, our proposed method splits the nodes of the original graph from coarse-grained to fine-grained, and generates the supernodes of the coarsened graph. Therefore, each granular-ball $\mathcal{GB}$ corresponds to a supernode in the coarsened graph $\mathcal{\overline{G}}$. Mathematically, 

\begin{equation}
\mathcal{\overline{V}}=\{\mathcal{GB}_1,\mathcal{GB}_2,\cdots,\mathcal{GB}_t\},t=\overline{N}, 
\end{equation}

\noindent which covers all nodes in $\mathcal{G}$. If there is already an edge between the nodes contained within two different granular-balls, then an edge is created between the two granular-balls. The process can be represented as follows:


\begin{equation}
\scalebox{0.9}{$
\mathcal{\overline{E}}\!=\!\{(\mathcal{GB}_i,\mathcal{GB}_j)\!\mid\!\exists e\!=\!(u,v)\!\in\!\mathcal{E},u\!\in\!\mathcal{\widetilde{V}}_{i},v\!\in\!\mathcal{\widetilde{V}}_{j},\mathcal{GB}_i\!\neq\!\mathcal{GB}_j\}.
$}
\end{equation}

This stage enhances the representation of graph by focusing on broader connectivity patterns, while maintaining the critical properties of the graph. 

In summary, the original graph is splitted from coarse-grained to fine-grained based on adaptive quality splitting condition, generating subgraphs with different granularity sizes. In this splitting process, the internal connection density within granular-ball is much higher than the connection density between granular-balls, thereby more effectively capturing the potential multi-granularity characteristics in the original graph structure, which is exactly what we aim to achieve. The quality of granular-ball reflects the inherent relationships and interactions within local areas of nodes, a characteristic that highly aligns with the learning requirements for local structural information in node classification tasks, and helps enhance the classification ability of the model.

\subsection{Theoretical analysis of GB-CGNN}
Due to the difference in the number of nodes between the original graph $\mathcal{G}$ and the coarsened graph $\mathcal{\overline{G}}$, we introduce appropriate intrinsic functionals such as the Rayleigh quotient (which is a quantity related to the structure of the graph and does not vary with the arrangement of nodes). The projection matrix $\mathbf{C}$, where $\mathbf{C}_{ij}=1$ if $v_i$ belongs to $\mathcal{GB}_j$, ensures that the spectral information of the two graphs becomes comparable \cite{cai2021graph}. Therefore, we first give the definitions of the Rayleigh quotient of the original graph and the coarsened graph respectively. Then the coarsening effect of GB-CGNN is proved by the Rayleigh quotient.

\textbf{Definition 2} For an original graph $\mathcal{G}$, its Rayleigh quotient is defined as:
\begin{equation}
R_o(\mathbf{\mathbf{L},x})=\frac{\mathbf{x}^T\mathbf{L}\mathbf{x}}{\mathbf{x}^T\mathbf{x}},
\end{equation}

\noindent where $\mathbf{L}$ is the Laplace matrix of $\mathcal{G}$, $\mathbf{x}$ is the eigenvector of $\mathbf{L}$. According to the definition of the projection matrix $\mathbf{C}$, the Rayleigh quotient of the coarsened graph obtained by GB-CGNN is given, which is defined as follows:

\textbf{Definition 3} For the coarsened graph $\mathcal{\overline{G}}$, its Rayleigh quotient is defined as:
\begin{equation}
R_c(\mathbf{\mathbf{\overline{L}},\mathbf{\overline{x}}})=\frac{(\mathbf{C}^T\mathbf{x})^T\mathbf{\overline{L}}(\mathbf{C}^T\mathbf{x})}{(\mathbf{C}^T\mathbf{x})^T(\mathbf{C}^T\mathbf{x})},
\end{equation}

\noindent where $\mathbf{\overline{L}}$ is the Laplace matrix of $\mathcal{\overline{G}}$. 

\textbf{Theorem 1} Assuming $\mathbf{C}$ is a generalized orthogonal matrix (i.e. $\mathbf{C}^T\mathbf{C}\approx\mathbf{I}$, where $\mathbf{I}$ is the identity matrix), we have:

\begin{equation}
R_c(\mathbf{\mathbf{\overline{L}},\mathbf{\overline{x}}})\approx R_o(\mathbf{\mathbf{L},x}).
\end{equation}

\begin{proof}
Considering the assumption that $\mathbf{C}$ is a generalized orthogonal matrix. For the denominator of $R_c(\mathbf{\mathbf{\overline{L}},\mathbf{\overline{x}}})$, using the orthogonality of $\mathbf{C}$, we get $(\mathbf{C}^T\mathbf{x})^T(\mathbf{C}^T\mathbf{x})\approx \mathbf{x}^T\mathbf{C}\mathbf{C}^T\mathbf{x}=\mathbf{x}^T\mathbf{I}\mathbf{x}=\mathbf{x}^T\mathbf{x}$. And then, for the numerator of $R_c(\mathbf{\mathbf{\overline{L}},\mathbf{\overline{x}}})$, we have $(\mathbf{C}^T\mathbf{x})^T\mathbf{\overline{L}}(\mathbf{C}^T\mathbf{x}) \approx \mathbf{x}^T\mathbf{C}\mathbf{\overline{L}}\mathbf{C}^T\mathbf{x}$. According to $\mathbf{\overline{L}}=\mathbf{C}^T\mathbf{L}\mathbf{C}$, we obtain $\mathbf{C}\mathbf{\overline{L}}\mathbf{C}^T \approx \mathbf{L}$. Thus, $\mathbf{x}^T\mathbf{C}\mathbf{\overline{L}}\mathbf{C}^T\mathbf{x}\approx\mathbf{x}^T\mathbf{L}\mathbf{x}$. Finally, using these approximations, the Rayleigh quotient of the coarsened graph obtained by GB-CGNN can be approximated as $R_c(\mathbf{\mathbf{\overline{L}},\mathbf{\overline{x}}})\approx R_o(\mathbf{\mathbf{L},x})$.
\end{proof}

\subsection{Training GCN with Granular-ball Coarsening Graph}
Next, we incorporate the coarsened graph $\overline{\mathcal{G}}$ into the GCN
models for training. Through the above granular-ball graph coarsening method, we obtain $\overline{N}$ subgraphs as follows:

\begin{equation}
\begin{split}
\overline{\mathcal{G}} &= \{\mathcal{GB}_1,\mathcal{GB}_2,\cdots,\mathcal{GB}_t\} \\
&= \{\mathcal{\widetilde{G}}_1,\mathcal{\widetilde{G}}_2,\cdots,\mathcal{\widetilde{G}}_t\} \\
&= \{(\mathcal{\widetilde{V}}_1,\mathcal{\widetilde{E}}_1),(\mathcal{\widetilde{V}}_2,\mathcal{\widetilde{E}}_2),\cdots,(\mathcal{\widetilde{V}}_t,\mathcal{\widetilde{E}}_t)\}, \\
&\quad t=\overline{N},
\end{split}
\end{equation}

\noindent where $\mathcal{\widetilde{E}}_i(i=1,2,\cdots,t)$ only consists of the edges between nodes in $\mathcal{\widetilde{V}}_i(i=1,2,\cdots,t)$. The adjacency matrix $\mathbf{A}$ becomes the following form:

\begin{equation}
\mathbf{A}=\mathbf{\widetilde{A}}+\mathbf{\Delta}=\begin{bmatrix}
A_{11} & \cdots & A_{1t} \\
\vdots & \ddots & \vdots \\
A_{t1} & \cdots & A_{tt}
\end{bmatrix},
\end{equation}

and

\begin{equation}
\mathbf{\widetilde{A}}=\begin{bmatrix}
A_{11} & \cdots & 0 \\
\vdots & \ddots & \vdots \\
0 & \cdots & A_{tt}
\end{bmatrix},\mathbf{\Delta}=\begin{bmatrix}
0 & \cdots & A_{1t} \\
\vdots & \ddots & \vdots \\
A_{t1} & \cdots & 0
\end{bmatrix},
\end{equation}

\noindent where each diagonal block $\mathbf{\widetilde{A}}_{ii}$ is a $\lvert \mathcal{\widetilde{V}}_i \rvert \times \lvert \mathcal{\widetilde{V}}_i \rvert$  adjacency matrix
containing the edges within $\mathcal{\widetilde{G}}_i$. $\mathbf{\widetilde{A}}_{ij}$ contains the edges between two edge sets $\mathcal{\widetilde{E}}_i$ and $\mathcal{\widetilde{E}}_j$. $\mathbf{\Delta}$ is
the matrix consisting of all off-diagonal blocks of $\mathbf{A}$. 

Similarly, we can partition the feature matrix $\mathbf{X}$ and training labels $\mathbf{Y}$ according to the granular-ball coarsening graph $\mathcal{GB}_1,\mathcal{GB}_2,\cdots,\mathcal{GB}_t$ as $\mathbf{X}_1,\mathbf{X}_2,\cdots,\mathbf{X}_t$ and $\mathbf{Y}_1,\mathbf{Y}_2,\cdots,\mathbf{Y}_t$, where
$\mathbf{X}_i(i=1,2,\cdots,t)$ and $\mathbf{Y}_i(i=1,2,\cdots,t)$ consist of the features and labels for the nodes in $\mathcal{GB}_i$, respectively. 

The benefit of this block-diagonal approximation $\overline{\mathcal{G}}$ is that the objective function of GCN becomes decomposible into different batches. Let $\mathbf{\widetilde{A}}'$ denotes the normalized version of $\mathbf{\widetilde{A}}$, the final embedding matrix becomes the following form due to the block-diagonal form of $\mathbf{\widetilde{A}}$ ($\mathbf{\widetilde{A}}'_{ii}$ is the corresponding diagonal block of $\mathbf{\widetilde{A}'}$):


\begin{equation}
\scalebox{0.80}{$
\begin{aligned}
\mathbf{H}^{(L)} &= \mathbf{\widetilde{A}}'\sigma(\mathbf{\widetilde{A}}'\sigma(\cdots \sigma(\mathbf{\widetilde{A}}'\mathbf{X}\mathbf{W}^{(0)})\mathbf{W}^{(1)})\cdots)\mathbf{W}^{(L-1)} \\
&= \begin{bmatrix}
\mathbf{\widetilde{A}}'_{11}\sigma(\mathbf{\widetilde{A}}'_{11}\sigma(\cdots \sigma(\mathbf{\widetilde{A}}'_{11}\mathbf{X}_{1}\mathbf{W}^{(0)})\mathbf{W}^{(1)})\cdots)\mathbf{W}^{(L-1)}\\
\vdots\\
\mathbf{\widetilde{A}}'_{tt}\sigma(\mathbf{\widetilde{A}}'_{tt}\sigma(\cdots \sigma(\mathbf{\widetilde{A}}'_{tt}\mathbf{X}_{t}\mathbf{W}^{(0)})\mathbf{W}^{(1)})\cdots)\mathbf{W}^{(L-1)}
\end{bmatrix}
\end{aligned}
$}
\label{1}
\end{equation}

Then, the loss can denoted as:

\begin{equation}
 \mathcal{L}_{\mathbf{\widetilde{A}}'_{ii}}=\frac{1}{\lvert\mathcal{\widetilde{V}}_i\rvert}\sum_{i \in \mathcal{\widetilde{V}}_i}loss(y_i,\hat{y}_i).
\end{equation}

We sample $p$ subgraphs (granular-balls) and then conduct SGD to update based on the gradient of $\mathcal{L}_{\mathbf{\widetilde{A}}'_{ii}}$. This only requires $\mathbf{A}_{ii}$, the $\mathbf{X}_i$
, $\mathbf{Y}_i$ of the subgraph $\mathcal{G}_i$ on the current batch. The implementation only necessitates the forward and backward propagation of matrix products (a single block from Formula (\ref{1})), which is significantly simpler to implement compared to the neighborhood search procedure employed in previous SGD-based training methods.

The total loss can be expressed as:
\begin{equation}
 \mathcal{L}_{\mathbf{\widetilde{A}'}} = \sum_{i=1}^{t}\mathcal{L}_{\mathbf{\widetilde{A}}'_{ii}}.
\label{lc}
\end{equation}

This method enhances the representation of the graph by focusing on a broader range of connection patterns, making the edge connections within granular-balls of different granularity sizes in the original graph tighter than those between different granular-balls. Sampling training is conducted on the coarsened subgraphs, which significantly reduces the computational cost. In terms of time complexity, the coarsening time of GB-CGNN is much lower than that of GBGC \cite{GBGC} and SGBGC \cite{xia2025graph}. The overall design of GB-CGNN is shown in Algorithm \ref{alg:GB-CGNN}.

\begin{algorithm} [h]
    \renewcommand{\algorithmicrequire}{\textbf{Input:}}
    \renewcommand{\algorithmicensure}{\textbf{Output:}}
    \caption{GB-CGNN}
    \label{alg:GB-CGNN}
    \begin{algorithmic}[1]
        \REQUIRE Original graph $\mathcal{G}$, Feature matrix $\mathbf{X}$, Labels $\mathbf{Y}$;
        \ENSURE Trained weight matrix $\mathbf{W}^*$
        \STATE The original graph $\mathcal{G}$ is regarded as the coarsest granular-ball;
        \STATE Generate initial granular-balls $\mathcal{GB}_{init}=\{\mathcal{GB}_1,\mathcal{GB}_2,\cdots,\mathcal{GB}_k\},k=\sqrt{N}$ by using the METIS algorithm;
        \STATE Calculate the quality of each granular-ball and decide whether to split initial granular-balls based on the adaptive splitting condition;
        \STATE Construct the granular-ball coarsening graph $\mathcal{\overline{G}} = \{\mathcal{GB}_1,\mathcal{GB}_2,\cdots,\mathcal{GB}_t,t=\overline{N}\}$ using granular-balls;
        \FOR {$iter=1,2,\cdots,max_{iter}$}
                \STATE Randomly choose $p$ granular-balls from $\mathcal{\overline{G}}$ without replacement;
                \STATE Form the subgraph 
                according to those selected granular-balls $\{\mathcal{GB}_1,\mathcal{GB}_2,\cdots,\mathcal{GB}_p\}$;
                \STATE Train parameters to minimize the loss $\mathcal{L}_{\mathbf{\widetilde{A}'}}$;
                \STATE Conduct Adam update using gradient estimate;
        \ENDFOR
        \STATE Return the trained weight matrix $\mathbf{W}^*$.
    \end{algorithmic}  
\end{algorithm}


The research work Cluster-GCN \cite{chiang2019cluster} demonstrates that this method maintains good node classification performance. The primary reason lies in the fact that its mini-batch does not randomly sample nodes, but constructs locally dense subgraphs based on graph clustering results. Furthermore, the training efficiency of mini-batch GCN is closely related to the ``embedding utilization". That is, if a batch contains more internal connections, the node representations can be reused by more neighbors, thereby reducing redundant computations and enhancing training efficiency. However, higher embedding utilization does not always lead to better node classification performance. Its effectiveness also depends on whether the batch subgraph can retain a compact and discriminative local structure. 

In contrast, GB-CGNN proposed in this paper adopts the granular-ball splitting strategy to adaptively construct training subgraphs in a coarse-to-fine manner, generating granular-ball subgraphs that are more compact, balanced, and have stronger local consistency. Moreover, the fitting ability of granular-ball is relatively strong, meaning that granular-ball can have a coarser granularity with fewer quantities in simple regions, and a finer granularity with more quantities in complex regions. Therefore, GB-CGNN not only enhances the quantity of embedding utilization but also improves its effectiveness, meaning that node representations are primarily reused within more meaningful local neighborhoods. This enables GB-CGNN to not only improve training efficiency but also retain more neighborhood discriminative information required for node classification, ultimately achieving better performance.

\subsection{Analysis of time and memory complexity}
\subsubsection{The time complexity of the first stage of GB-CGNN}

The time complexity of this stage is $O(N)$, and the derivation is as follows. 


\textbf{The coarse-grained initialization.} The METIS algorithm mainly consists of three stages: the coarsening stage, the initial partitioning stage, and the uncoarsening and refinement stage. 

For an original graph $\mathcal{G}$ with $N$ nodes and $E$ edges, in the first stage, the matching algorithm traverses the adjacent edges of each node and selects the edge with the largest weight for matching. The time complexity of this algorithm is linearly related to the number of edges at the current level. Ideally, the number of nodes is reduced by approximately half after each coarsening. Therefore, the time complexity of the entire coarsening stage is the sum of the time complexities of all levels, that is, $O(E)$. 

In the second stage, an initial partition is performed on the smallest coarsened graph. Since the number of partitions is specified, the time complexity of this stage is $O(1)$.

In the third stage, at each step, the Fiduccia-Mattheyses (FM) algorithm or Kernighan-Lin (KL) algorithm is used for boundary optimization to reduce the number of cut edges. Firstly, restore the partitioning result to the original graph, and directly assign the partitioning scheme of the current level's coarsened graph to the previous level's coarsened graph. The time complexity of the uncoarsening stage is the sum of the time complexities of all levels. Therefore, the time complexity of uncoarsening is $O(N)$. Secondly, the boundary optimization algorithm FM only focuses on the vertices near the cutting edge. Similarly, the time complexity of the refinement stage is the sum of the time complexities of each level. Therefore, the optimized time complexity is $O(E)$. Therefore, the total time complexity of this stage is $O(N+E)$. Since most graphs are sparse graphs (i.e., the $AD$ of each vertex is a very small constant), and the number of edges is proportional to the number of vertices, the total time complexity of the METIS algorithm can be simplified to $O(N)$. Thus, the time complexity of the coarse-grained initialization of GB-CGNN is $O(N)$.

\textbf{The fine-grained binary splitting.} For a granular-ball $\mathcal{GB}$ has $\widetilde{N}$ nodes and $\widetilde{E}$ edges, it involves finding the node with the highest degree, which requires a time complexity of $\widetilde{N}$. Each center node traverses the remaining nodes layer by layer through BFS, and other non-center nodes are assigned to the center node according to who was first traversed by a center node's BFS. Therefore, the time complexity of each BFS is $O(\widetilde{N}+\widetilde{E})$. Since it is a binary splitting, we need to do BFS traversal for each center node, and the total complexity is $O(2(\widetilde{N}+\widetilde{E}))$. Since each non-center node only needs to be accessed the first time in a BFS process, this process can be done simultaneously with BFS traversal without additional complexity. Therefore, the time complexity of the fine-grained binary splitting is $O(\widetilde{N}+2(\widetilde{N}+\widetilde{E}))$. In summary, since $\widetilde{N} < N$ and $\widetilde{E} < E$, the time complexity of \textit{the adaptive granular-ball coarsening graph formation} is $O(N)$. 

\subsubsection{The time complexity of GB-CGNN}

The time complexity of GB-CGNN is $O(L||\mathbf{A}||_0F+LNF^2)$, and the derivation is as follows. 


Since each node in $\mathcal{GB}_i$ only links to nodes inside $\mathcal{GB}_i$, each node does not need to perform neighborhoods searching outside $\mathbf{\widetilde{A}}_{ii}$. The calculation of each batch only involves the product of some element-wise
operations, so the time complexity of each batch is $O(||\mathbf{\widetilde{A}}_{ii}||_0F+BF^2)$, where B denotes the batch size. Therefore, the time complexity of each epoch is $O(||\mathbf{A}||_0F+NF^2)$, where $||\mathbf{A}||_0$ denotes the number of nonzeros in the adjacency matrix. For the $L$-layer GCN network, the time complexity at this stage is $O(L||\mathbf{A}||_0F+LNF^2)$. 

\subsubsection{The memory complexity of GB-CGNN}

The memory complexity of GB-CGNN is $O(BLF+LF^2)$, and the derivation is as follows. 

Our GB-CGNN only needs to load subgraphs rather than the entire graph into GPU memory. That is to say, our GB-CGNN only needs to load $B$ samples in each batch and store their embeddings on each layer. Therefore, for the $L$-layer GCN network, the memory for storing embeddings is $O(BLF)$, and the memory for storing $\{\mathbf{W}^{(l)}\}_{l=1}^{L}$ is $O(LF^2)$. For simplicity, we omit the memory used to store graphs or subgraphs here because they are fixed.

\section{Experiments}

In this section, we conduct a series of experiments aimed at assessing the performance of our proposed GB-CGNN. The experiments are structured to answer the following several research questions (RQs):
\begin{itemize}
  \item \textbf{RQ1:} Is the proposed GB-CGNN superior to existing SOTA methods in terms of efficiency and performance of graph coarsening and GCN training?
  \item \textbf{RQ2:} What is the impact of the critical components of GB-CGNN on its overall performance? 
  \item \textbf{RQ3:} How do the hyperparameters of GB-CGNN affect its performance?
  \item \textbf{RQ4:} How does GB-CGNN perform in specific case studies?
  \item \textbf{RQ5:} How do different quality evaluations affect the coarsening time and performance of GB-CGNN?
\end{itemize}

\subsection{Experimental settings}
\textbf{Datasets.} We evaluated the performance of our proposed GB-CGNN and some comparison GCN training algorithms on several node classification datasets. Details of datasets are shown in Table \ref{datasettable}. Among them, the settings of the datasets PPI, Reddit, Flickr, Yelp, and Amazon are consistent with those in reference \cite{zenggraphsaint}. The remaining datasets are segmented using 60\%/ 20\%/ 20\% random segmentation for training, validation and testing. 

\begin{table}[tb!]
    \centering
    \caption{Summary of information from the benchmark datasets (“m” stands for multi-label, and “s” for single-label).}
    \label{datasettable}
    \begin{tabularx}{\linewidth}{@{} *{5}{>{\centering\arraybackslash}X} @{}} 
        \toprule
        \textbf{Datasets}  & \textbf{Nodes}  & \textbf{Edges} & \textbf{Features} & \textbf{Classes} \\
        \midrule
        Cora &  2,708 & 5,429 & 1433 & 7 (s)  \\
        Citeseer &  3,327 & 4,732 & 3703 & 6 (s)  \\
        PubMed &  19,717 & 44,338 & 500 & 3 (s)  \\
        \hline
        Photo  &  7,650 & 119,081 & 745 & 8 (s)  \\
        Computers &  13,752 & 245,861 & 767 & 10 (s)  \\
        Co-Cs &  18,333 & 182,121 & 6805 & 15 (s)  \\
        Co-phy &  34,493 & 247,962 & 8415 & 5 (s)  \\
        PPI &  56,944 & 818,716 & 50 & 121 (m) \\
        Flickr &  89,250 & 899,756 & 500 & 7 (s)  \\
        \hline
        Reddit &  232,965 & 11,606,919 & 602 & 41 (s)  \\
        Yelp  &  716,847 & 6,977,410 & 300 & 100 (m) \\
        Amazon  &  1,598,960 & 818,716 & 200 & 107 (m) \\
        \bottomrule
    \end{tabularx}
\end{table}
\textbf{Comparison baselines.} To demonstrate the effectiveness of our proposed GB-CGNN, we compared GB-CGNN with baselines from the following two aspects. 
\begin{itemize}
  \item \textbf{Graph coarsening methods:} (1) SGBGC \cite{xia2025graph}. (2) GBGC \cite{GBGC}. (3) SCAL \cite{huang2021scaling}. (4) FGC \cite{kumar2023featured}. (5) JCGC \cite{ron2011relaxation}. (6) GSGC \cite{ron2011relaxation}. (7) VNGC \cite{loukas2019graph}. (8) VEGC \cite{loukas2019graph}.
  \item \textbf{GNN methods:} (1) GCN \cite{kipf2016semi}. (2) GAT \cite{velivckovic2017graph}. (3) GraphSAGE \cite{hamilton2017inductive}. (4) FastGCN \cite{chen2018fastgcn}. (5) AS-GCN \cite{huang2018adaptive}. (6) Cluster-GCN \cite{chiang2019cluster}. (7) GraphSAINT \cite{zenggraphsaint}. (8) SGBGC \cite{xia2025graph}.
\end{itemize} 

\textbf{Performance metrics.} In our experimental implementation, based on the comparison methods of graph coarsening and GNN, a total of two performance evaluation metrics were set, namely coarsening time and $Micro\_F1$ score. 
\begin{itemize}
  \item \textbf{For graph coarsening methods,} the main comparison is the running time of the coarsening process.
  \item \textbf{For GNN methods,} similar to previous studies, we mainly employ $Micro\_F1$ score as the metric to measure the performance of node classification on multi-label datasets (PPI, Yelp, and Amazon). The definition of $Micro\_F1$ is as follows:

\begin{equation}
Micro\_F1=2\times\frac{Precision\times Recall}{Precision+Recall},
\end{equation}

\noindent where $Precision=\frac{TP}{TP+FP}$ and $Recall=\frac{TP}{TP+FN}$. $TP$, $FP$, $FN$ denote true positive, false positive, and false negative, respectively. $Micro\_F1$ score does not need to distinguish categories, and directly uses the $Precision$ and $Recall$ of the overall sample to calculate the $Micro\_F1$ score. 

For single-label datasets, we primarily use accuracy ($ACC$, which is equivalent to the $Micro\_F1$ score) as the metric to measure the performance of node classification. The definition of $ACC$ is as follows:

\begin{equation}
ACC=\frac{1}{N_t}\sum_{i=1}^{N_t}\mathds{1}(y_i=\hat{y}_i),
\end{equation}

\noindent where $N_t$ represents the total number of test nodes, $y_i$ and $\hat{y}_i$ represent the true label and predicted label of the $i$-th node, respectively. $\mathds{1}(\cdot)$ is an indicator function that takes the value of 1 when the condition is met, and 0 otherwise.

\end{itemize} 
\textbf{Implementation details.} The experiments are conducted on the Intel(R) Xeon(R) W-2245 CPU @ 3.90GHz with NVIDIA GeForce RTX 3090. The key software environment includes Python 3.8.20, Tensorflow-gpu 2.2.0, and NetworkX 1.11. In addition, we use grid search to adjust the hyperparameters and find the best combination. In GB-CGNN, we limit the number of convolutional layer ($L$) to the range \{2, 3, 4, 5, 7\} and search for the dimension of hidden layer ($H$) in the range \{128, 512, 1024, 2048\}. The dropout rate ($D$) is selected from \{0.2, 0.5\}. To ensure fairness of comparison, for a given dataset, we keep hyperparameters the same across all methods. For all methods we use the Adam optimizer with learning rate as 0.01, and weight decay as 0. For AS-GCN and GraphSAGE, we follow the settings in the papers and set the batch size ($B$) as 256 and 512, respectively. For Cluster-GCN, the number of subgraphs ($t$) is consistent with our GB-CGNN.  We use the default settings of sampled sizes for each layer (S1=25, S2=10) and the number of training epochs in GraphSAGE. 


\subsection{Performance of graph coarsening and node classification \textbf{(RQ1)}}

\textbf{Running time for graph coarsening methods.} To answer \textbf{RQ1}, we first compared the proposed GB-CGNN with some graph coarsening methods regarding the running time for coarsening. MGC \cite{jin2020graph}, SGC \cite{jin2020graph} and KGC \cite{chen2023gromov} are algorithms for graph classification tasks (with few nodes and edges). For datasets like Cora, they all fail to produce results due to memory overload. Therefore, we have not compared them with MGC, SGC and KGC here. The specific comparison results are shown in Table \ref{coarseningtimetable}, and we can observe the following:

\begin{itemize}
\item The coarsening runtime of GB-CGNN has consistently outperformed that of the comparison graph coarsening methods across all datasets, particularly on large-scale graph datasets such as Reddit, Yelp, and Amazon, where the advantages of GB-CGNN become even more pronounced (with coarsening times increasing by several orders of magnitude). In contrast, some graph coarsening methods are unable to even execute on large-scale graph datasets due to their computational inefficiencies.

\item The coarsening efficiency of GB-CGNN is exceptionally high, which can be attributed to its distinctive coarse-grained initial granular-ball generation and fine-grained binary splitting strategy. Specifically, these strategies first employ the efficient graph partitioning algorithm METIS for coarse-grained granular-ball generation, providing an effective initialization for the subsequent binary splitting process. Subsequently, fine-grained binary splitting is performed through local computations, ensuring an overall linear time complexity. Through granular-ball graph coarsening, GB-CGNN effectively captures the multi-granularity structural features of the original graph.

\end{itemize}

\begin{table*}[tb!]
    \centering
    \caption{Comparison of the running time for coarsening in graph coarsening methods (Unit: Second), where the best results are bolded. Results not reported are due to out-of-memory or unavailable. GB: GB-CGNN.}
    \label{coarseningtimetable}
    \begin{tabularx}{\linewidth}{@{} >{\centering\arraybackslash}X
        !{\vrule width 0.8pt}
        *{3}{>{\centering\arraybackslash}X}
        !{\vrule width 0.8pt}
        *{6}{>{\centering\arraybackslash}X}
        !{\vrule width 0.8pt}
        *{3}{>{\centering\arraybackslash}X} @{}}
        \toprule
        \textbf{Methods}
        & \textbf{Cora} & \textbf{Citeseer} & \textbf{PubMed}
        & \textbf{Photo} & \textbf{Computers} & \textbf{Co-Cs} & \textbf{Co-phy} & \textbf{PPI} & \textbf{Flickr}
        & \textbf{Reddit} & \textbf{Yelp} & \textbf{Amazon} \\
        \midrule
        SGBGC
        & 1.07 & 0.91 & 42.10
        & 23.32 & 44.44 & 44.70 & 92.75 & 59.03 & 625.63
        & 30329.22 & 74421.28 & - \\
        
        GBGC
        & 3.70 & 4.28 & 51.53
        & 66.83 & 225.51 & 58.21 & 205.74 & 561.76 & 1453.01
        & - & - & - \\
        
        SCAL
        & 2.58 & 3.41 & 18.45
        & 19.00 & 35.01 & 32.29 & 275.54 & 124.06 & 130.20
        & 2769.17 & - & - \\
        
        FGC
        & 7.19 & 14.37 & 314.87
        & 222.08 & 1711.53 & 879.99 & 4487.78 & - & -
        & 2062.22 & - & - \\
        
        JCGC
        & 8.59 & 9.34 & 169.26
        & 25.71 & 361.94 & 301.98 & 1357.26 & - & 2895.46
        & - & - & - \\
        
        GSGC
        & 10.40 & 10.10 & 1797.42
        & 34.95 & 1634.49 & 1203.65 & 6356.16 & - & 13075.61
        & - & - & - \\
        
        VNGC
        & 10.03 & 10.25 & 125.04
        & 36.09 & 43.55 & 218.10 & 465.96 & - & 348.38
        & 1045.13 & - & - \\
        
        VEGC
        & 9.04 & 9.58 & 162.66
        & 27.15 & 39.33 & 284.64 & 1201.92 & - & 314.63
        & 943.88 & - & - \\
        
        GB
        & \textbf{0.07} & \textbf{0.04} & \textbf{0.70}
        & \textbf{0.39} & \textbf{0.74} & \textbf{0.69} & \textbf{1.80} & \textbf{8.69} & \textbf{4.54}
        & \textbf{121.56} & \textbf{307.18} & \textbf{4791.81} \\
        \hline
        -w/o B
        & 0.04 & 0.03 & 0.35
        & 0.26 & 0.55 & 0.42 & 1.33 & 7.83 & 3.73
        & 109.81 & 255.88 & 5313.28 \\
        \hline
        -w/o I
        & 0.02 & 0.01 & 0.09
        & 0.10 & 0.15 & 0.08 & 0.33 & 0.49 & 0.49
        & 14.44 & 11.83 & 396.64 \\
        \bottomrule
    \end{tabularx}
\end{table*}

\begin{table*}[tb!]
    \centering
    \caption{The result of $ACC$ for node classification on single-label datasets. The results not reported are due to out-of-memory or unavailable. Cluster: Cluster-GCN.}
    \label{acctable}
    \begin{tabularx}{\linewidth}{@{} >{\centering\arraybackslash}X
        !{\vrule width 0.8pt}
        *{3}{>{\centering\arraybackslash}X}
        !{\vrule width 0.8pt}
        *{5}{>{\centering\arraybackslash}X}
        !{\vrule width 0.8pt}
        *{1}{>{\centering\arraybackslash}X} @{}}
        \toprule
        \textbf{Methods}
        & \textbf{Cora} & \textbf{Citeseer} & \textbf{PubMed}
        & \textbf{Photo} & \textbf{Computers} & \textbf{Co-Cs} & \textbf{Co-phy} & \textbf{Flickr}
        & \textbf{Reddit} \\
        \midrule
        GCN
        & 0.8577 & 0.7368 & 0.8619 & 0.9242 & 0.8651 & 0.6030 & 0.9565 & 0.4197 & 0.9330 \\
        
        GAT
        & 0.8637 & 0.7300 & 0.8780  & 0.8200 & 0.8093 & 0.9231 & 0.9231 & - & - \\
        
        GraphSAGE
        & 0.8017 & 0.6385 & 0.8777
        & 0.9266 & 0.8230 & 0.9252 & 0.9564 & 0.3992
        & 0.9479 \\
        
        FastGCN
        & 0.8408 & \textbf{0.7640} & 0.8692
        & 0.8865 & 0.8970 & 0.9010 & 0.8960 & 0.4207
        & 0.9240 \\
        
        AS-GCN
        & 0.8037 & 0.6771 & 0.8486
        & - & - & - & - & 0.4197
        & 0.9580 \\
        
        Cluster
        & 0.8513 & 0.6879 & 0.8809
        & 0.9467 & 0.8958 & 0.9287 & 0.9581 & 0.4553
        & 0.9625 \\
        
        GraphSAINT
        & 0.8561 & 0.7172 & 0.8807
        & 0.9396 & 0.8948 & 0.9245 & 0.9548 & 0.4829
        & 0.9610 \\
        
        SGBGC
        & 0.8572 & 0.6832 & 0.8301
        & 0.6577 & 0.7023 & 0.9263 & 0.9517  & -
        & - \\
        
        GB-CGNN
        & \textbf{0.8667} & 0.6306 & \textbf{0.8813}
        & \textbf{0.9537} & \textbf{0.9001} & \textbf{0.9331} & \textbf{0.9584} & \textbf{0.4845}
        & \textbf{0.9642} \\
        \hline
        -w/o B
        & 0.8571 & 0.6899 & 0.8795
        & 0.9510 & 0.9064 & 0.9309 & 0.9575 & 0.4831 & 0.9626 \\
        \hline
        -w/o I
        & 0.5442 & 0.5115 & 0.8780
        & 0.9342 & 0.9101 & 0.8958 & 0.9582 & -
        & - \\
        \bottomrule
    \end{tabularx}
\end{table*}

\begin{table}[tb!]
    \centering
    \caption{The result of $Micro\_F1$ score for node classification on multi-label datasets. The results not reported are due to out-of-memory or unavailable.}
    \label{F1table}
    \begin{tabular}{@{} l *{3}{c} @{}}
        \toprule
        \textbf{Methods} & \textbf{PPI} & \textbf{Yelp} & \textbf{Amazon} \\
        \midrule
        GCN          & 0.4608 & 0.1741 & 0.2810 \\
        GAT          & 0.9730 & 0.6122 & 0.4750 \\
        GraphSAGE         & 0.5761 & 0.6163 & 0.4842 \\
        FastGCN      & 0.4608 & 0.2651 & 0.5490 \\
        AS-GCN           & 0.6623 & -      & -      \\
        Cluster-GCN      & 0.9903 & 0.6030 & 0.7121 \\
        GraphSAINT        & 0.9038 & -      & -      \\
        SGBGC        & -      & -      & -      \\
        GB-CGNN           & \textbf{0.9909} & \textbf{0.6268} & \textbf{0.7394} \\
        \hline
        -w/o B       & 0.9901 & 0.6129 & -      \\
        -w/o I       & 0.7612 & -      & -      \\
        \bottomrule
    \end{tabular}
\end{table}

\textbf{Performance for node classification.} Node classification tasks are typically used to evaluate the quality of learned node embeddings. In this subsection, we mainly evaluate the effectiveness of GB-CGNN in node classification on twelve datasets. We use $ACC$ and $Micro\_F1$ score as evaluation metric for node classification performance. The types of datasets include citation networks, social networks, biological interaction networks, and commodity purchase networks, etc. Table \ref{acctable} and Table \ref{F1table} list the detailed results of $ACC$ and $Micro\_F1$ score of all methods on these datasets, respectively. In each dataset, the best results are highlighted in bold, while unreported results are out of memory or unavailable. The experimental results show that, the performance of GB-CGNN on most datasets is superior to all baseline methods. Specifically, our method achieved the best performance on 11/12 datasets, outperforming other baseline methods. For instance, on relatively large-scale graph datasets such as Yelp and Amazon, due to memory limitations, some baselines have no results. The performance of GB-CGNN is 1.70\% and 3.83\% higher than that of the second-best method, respectively. On the Photo dataset, its $Micro\_F1$ score outperforms the second-best method by over 0.74\%, and on the remaining datasets, the average relative improvement rate is 0.25\%. This observation demonstrate that the ability of GB-CGNN to handle information of different granularities is a key factor for its excellent performance in the node classification task. 

Overall, the advantage of GB-CGNN lies in its ability to fully consider the multi-granularity information of the graph structure. The granular-ball splitting mechanism can adaptively refine the graph structure from coarse-grained to fine-grained, continuously dividing regions with high internal heterogeneity or loose structure, thereby generating more compact, uniform, and locally homogeneous subgraphs. Such subgraphs not only reduce the information loss caused by the unreasonable removal of key adjacency relationships, but also align with the mechanism of GCN relying on local neighborhood aggregation, thus more effectively preserving the discriminative information required for node classification and ultimately obtaining superior vector representations. This further demonstrates the superiority of the algorithm in processing graph data with complex structures.

\begin{figure*}[tb!]
    \centering

    \subfigure[Cora]{
        \includegraphics[width=0.22\textwidth]{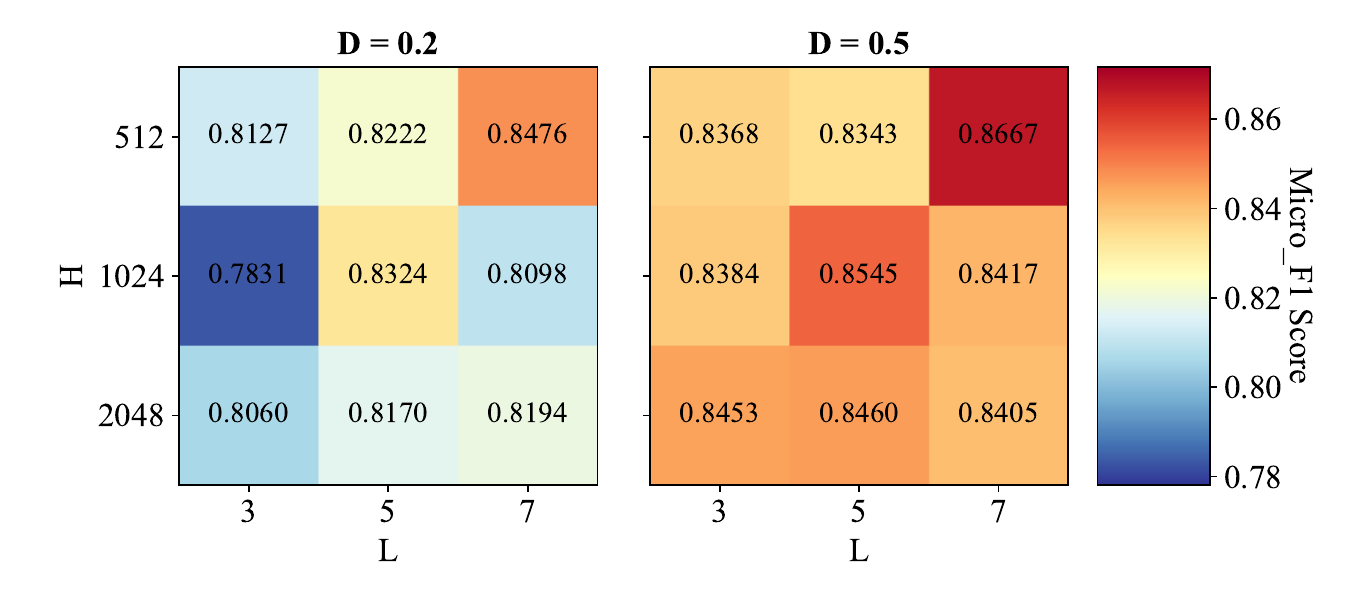}
    }
    \subfigure[Citeseer]{
        \includegraphics[width=0.22\textwidth]{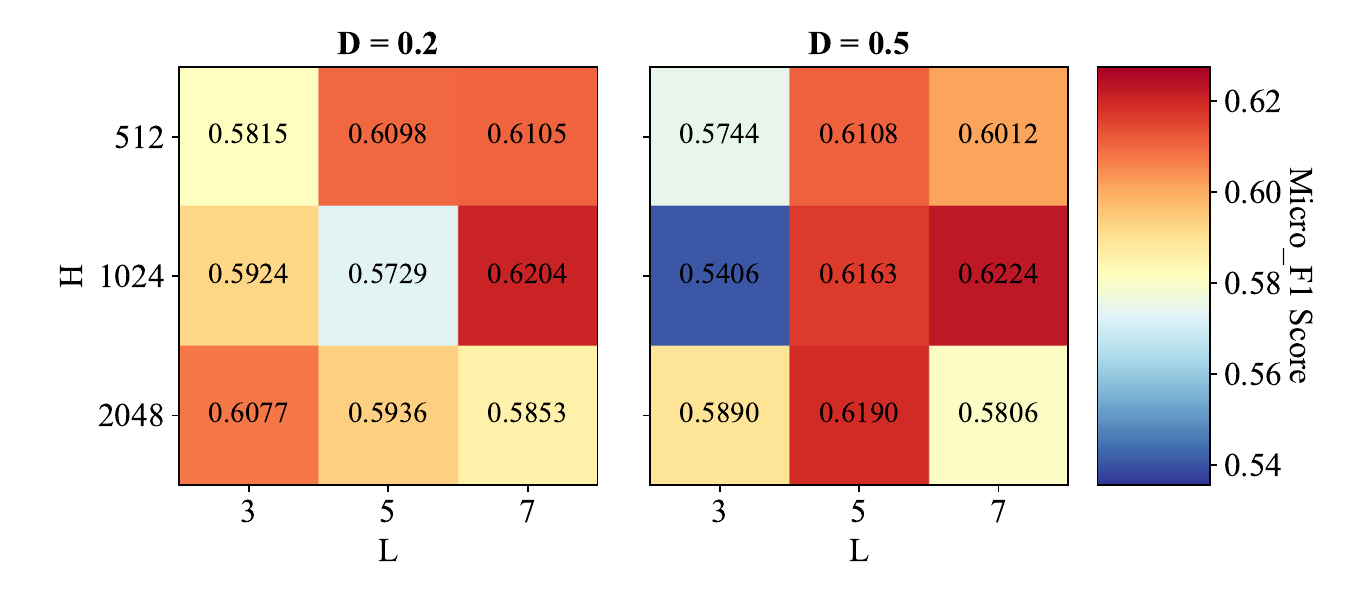}
    }
    \subfigure[PubMed]{
        \includegraphics[width=0.22\textwidth]{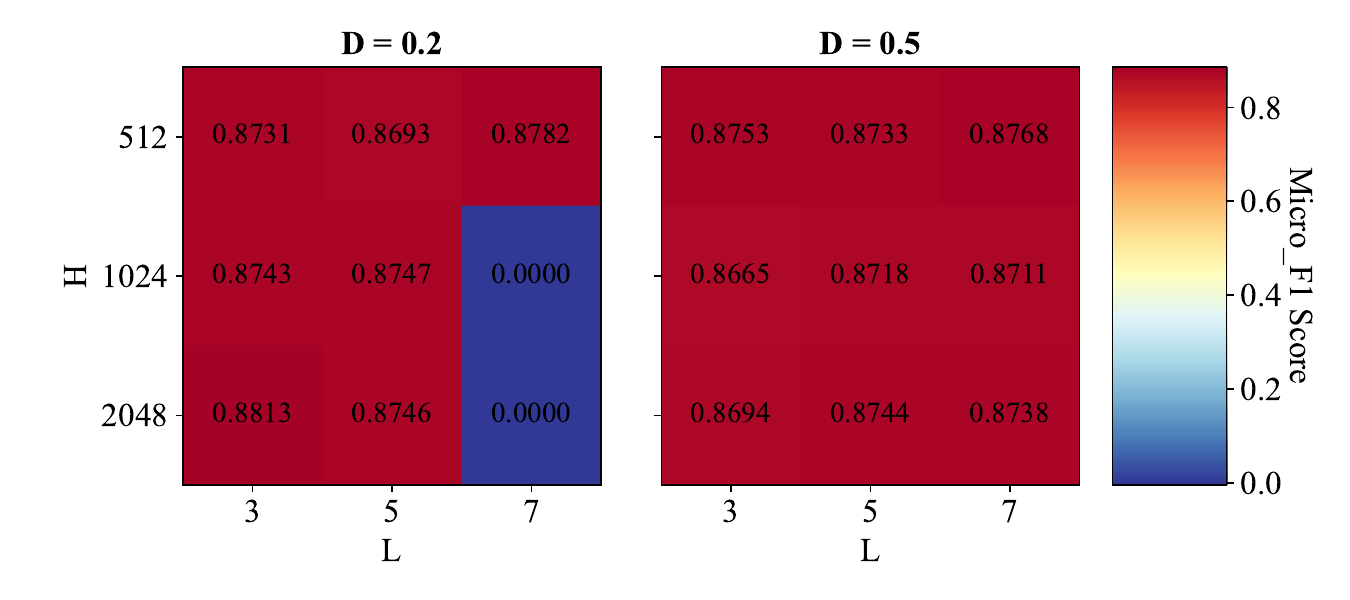}
    }
    \subfigure[Co-Cs]{
        \includegraphics[width=0.22\textwidth]{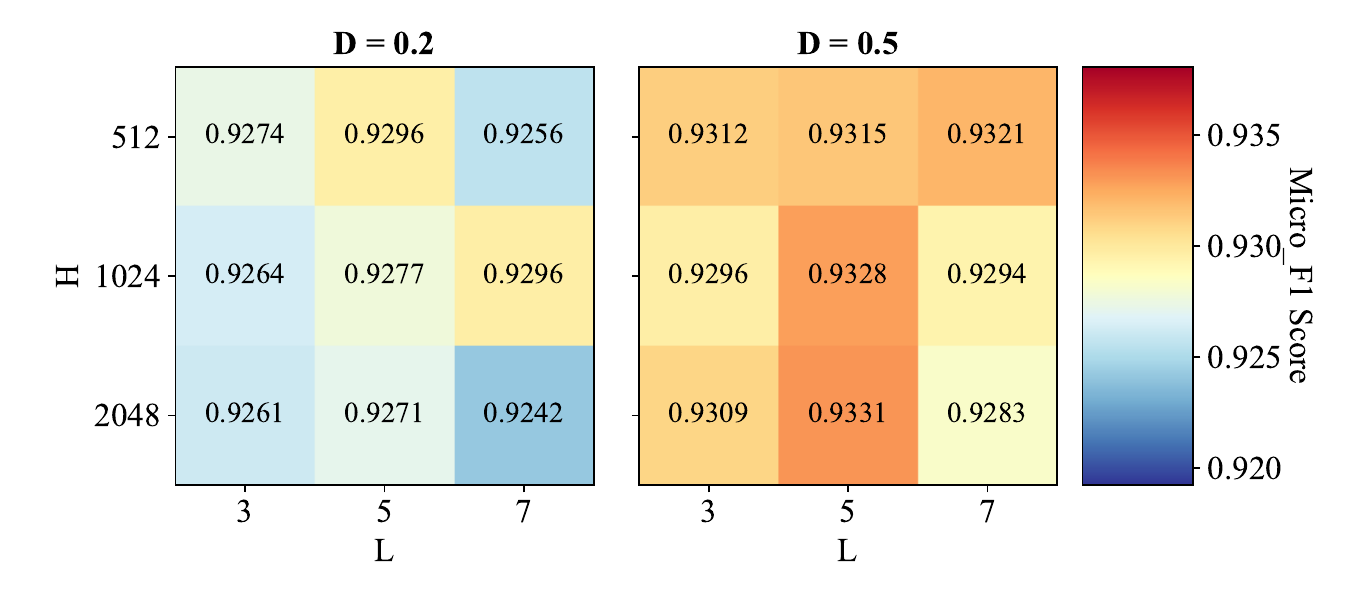}
    }

    \subfigure[Co-phy]{
        \includegraphics[width=0.22\textwidth]{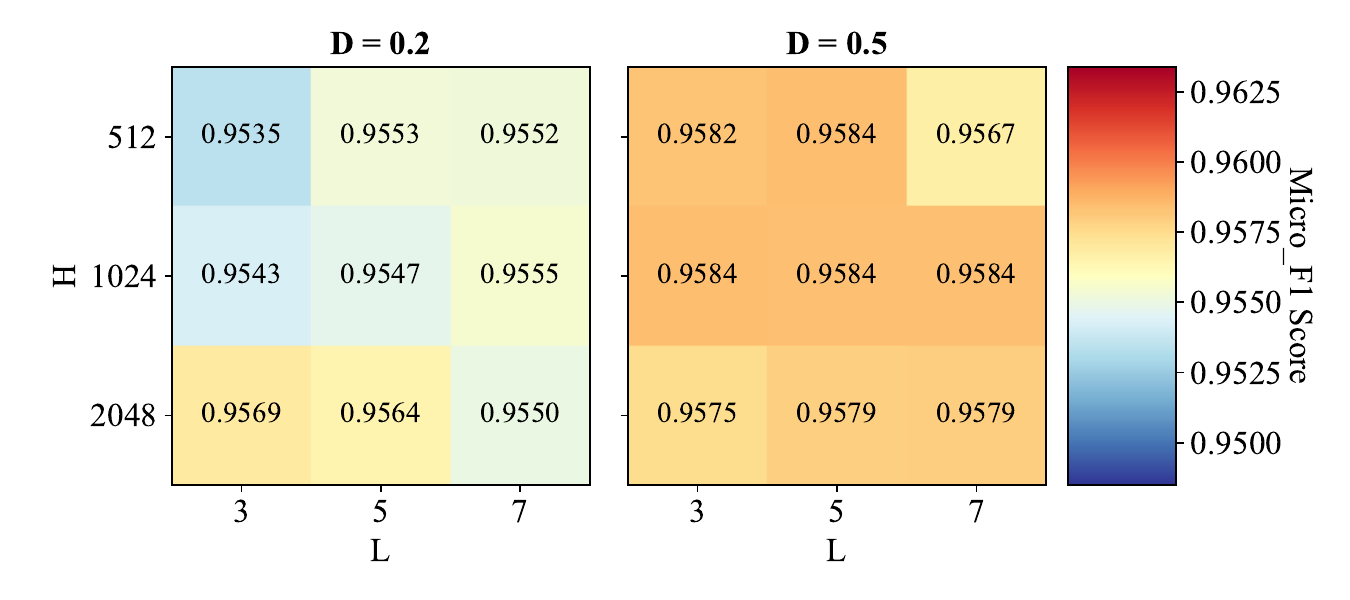}
    }
    \subfigure[PPI]{
        \includegraphics[width=0.22\textwidth]{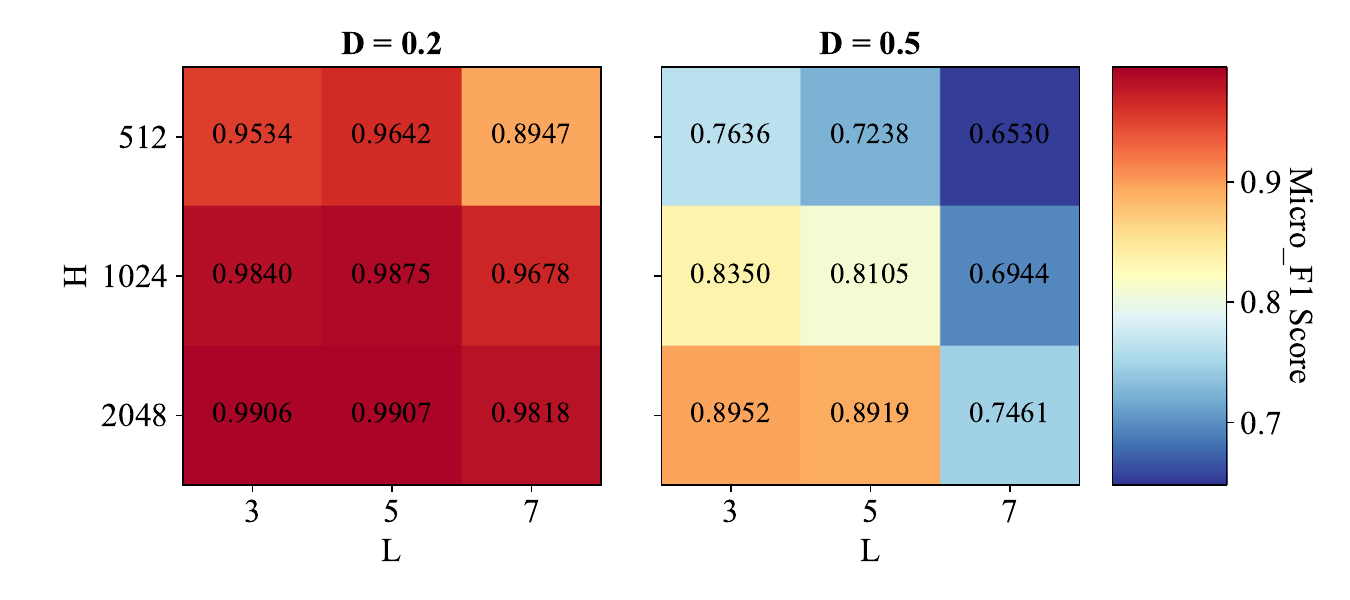}
    }
    \subfigure[Reddit]{
        \includegraphics[width=0.22\textwidth]{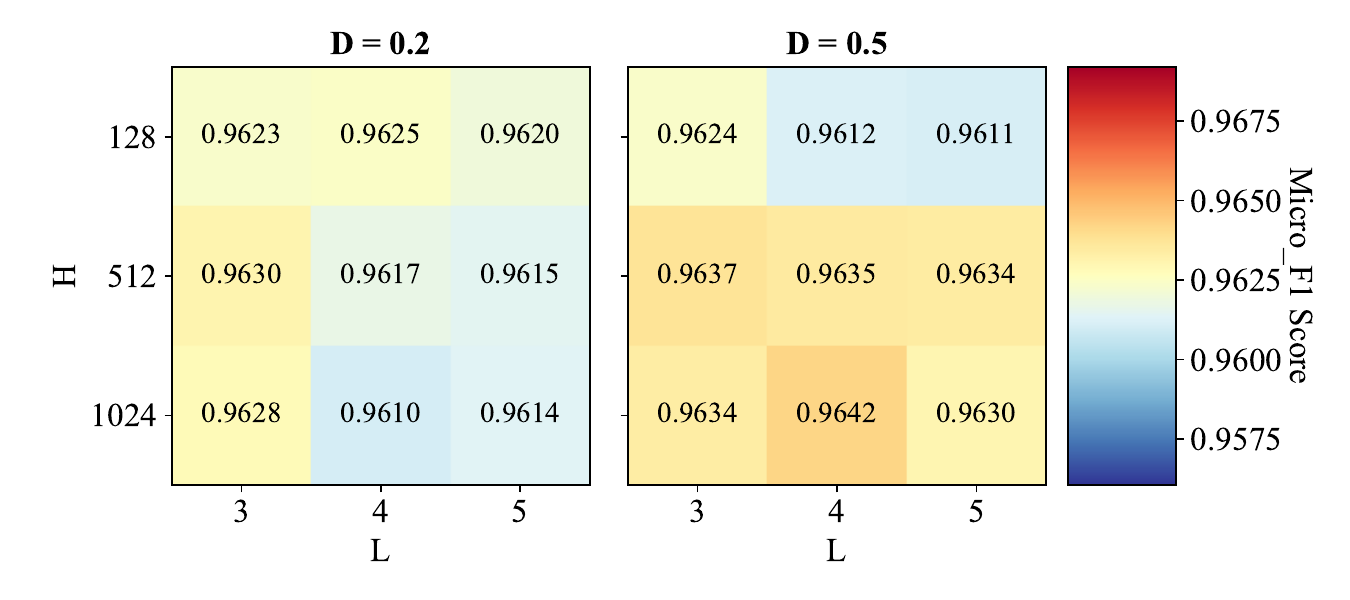}
    }
    \subfigure[Flickr]{
        \includegraphics[width=0.22\textwidth]{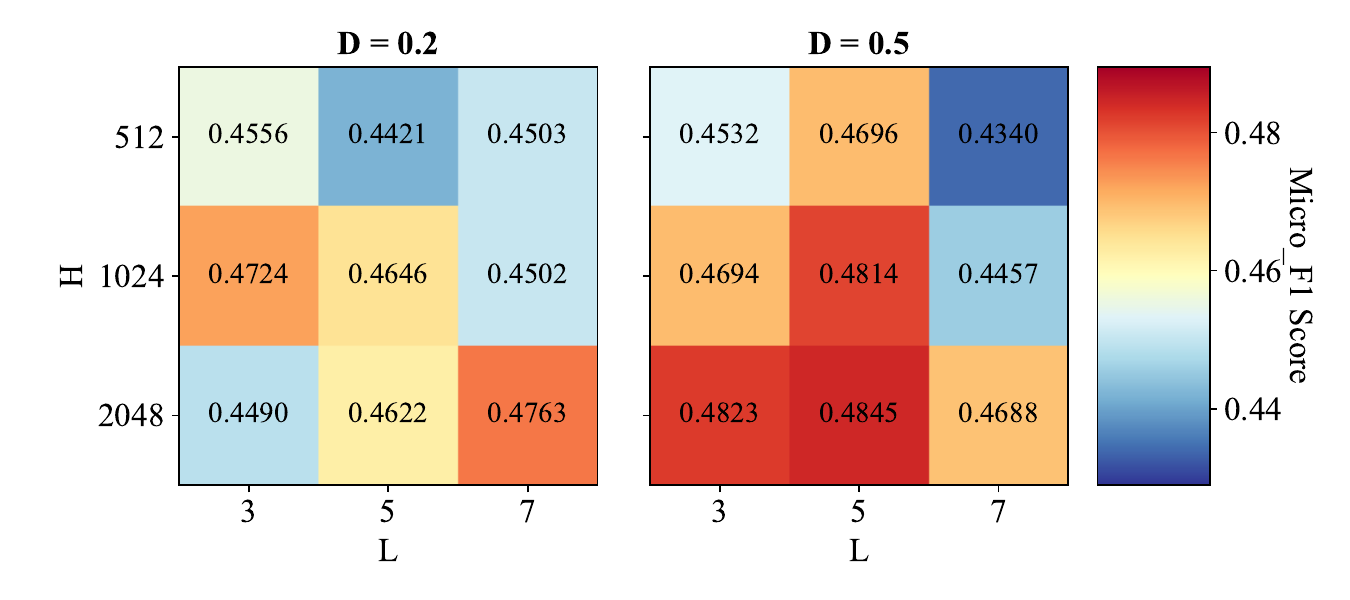}
    }

    \subfigure[Computers]{
        \includegraphics[width=0.22\textwidth]{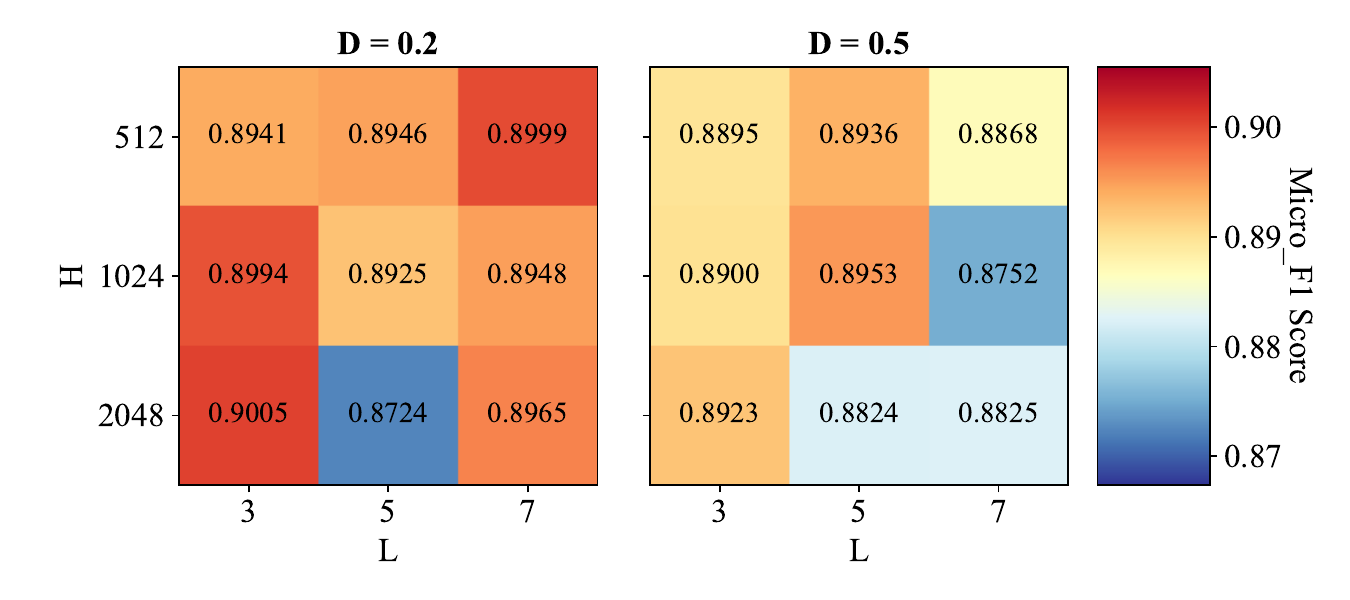}
    }
    \subfigure[Photo]{
        \includegraphics[width=0.22\textwidth]{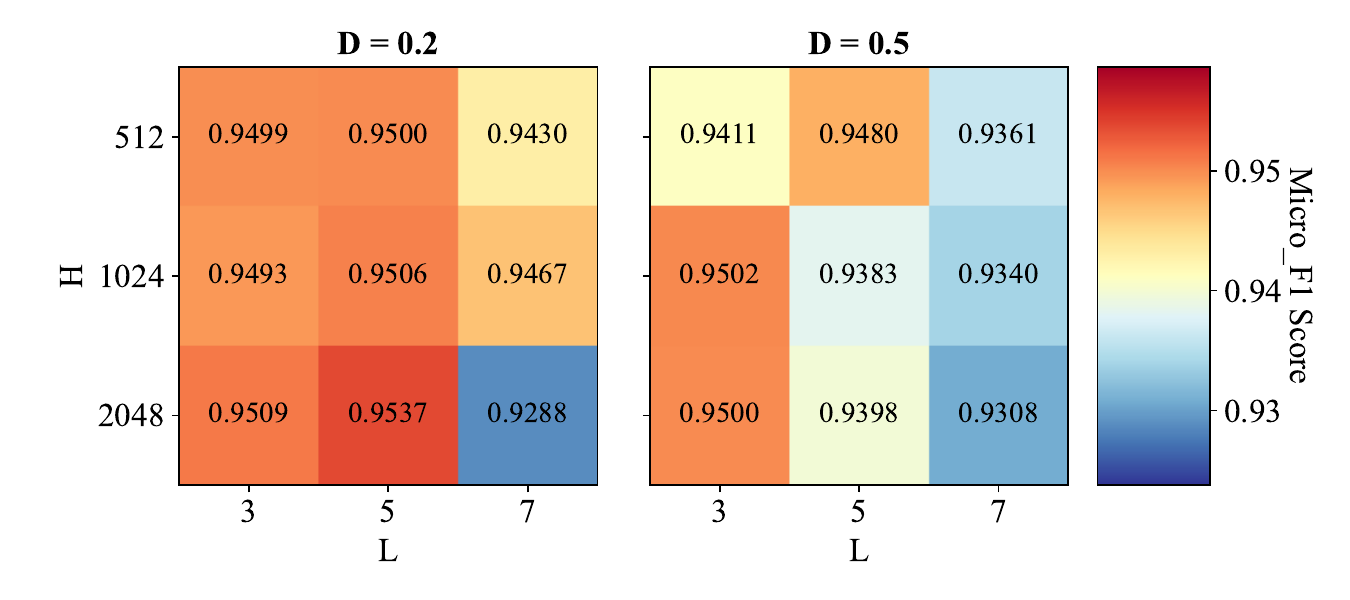}
    }
    \subfigure[Yelp]{
        \includegraphics[width=0.22\textwidth]{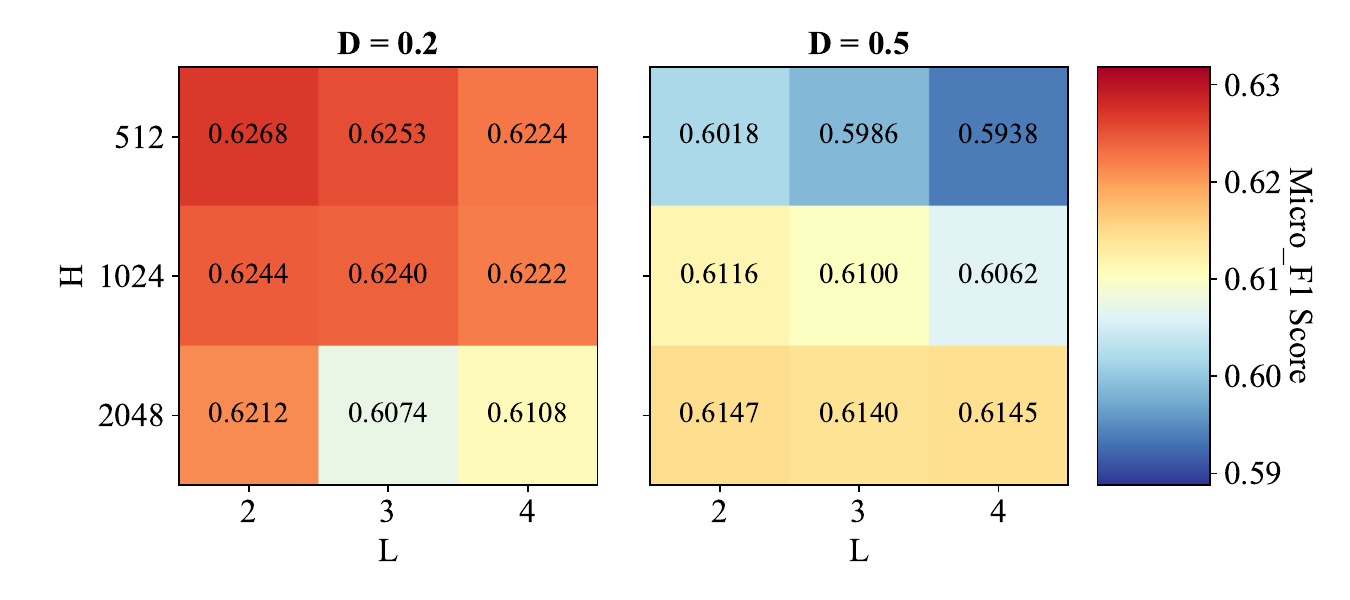}
    }
    \subfigure[Amazon]{
        \includegraphics[width=0.22\textwidth]{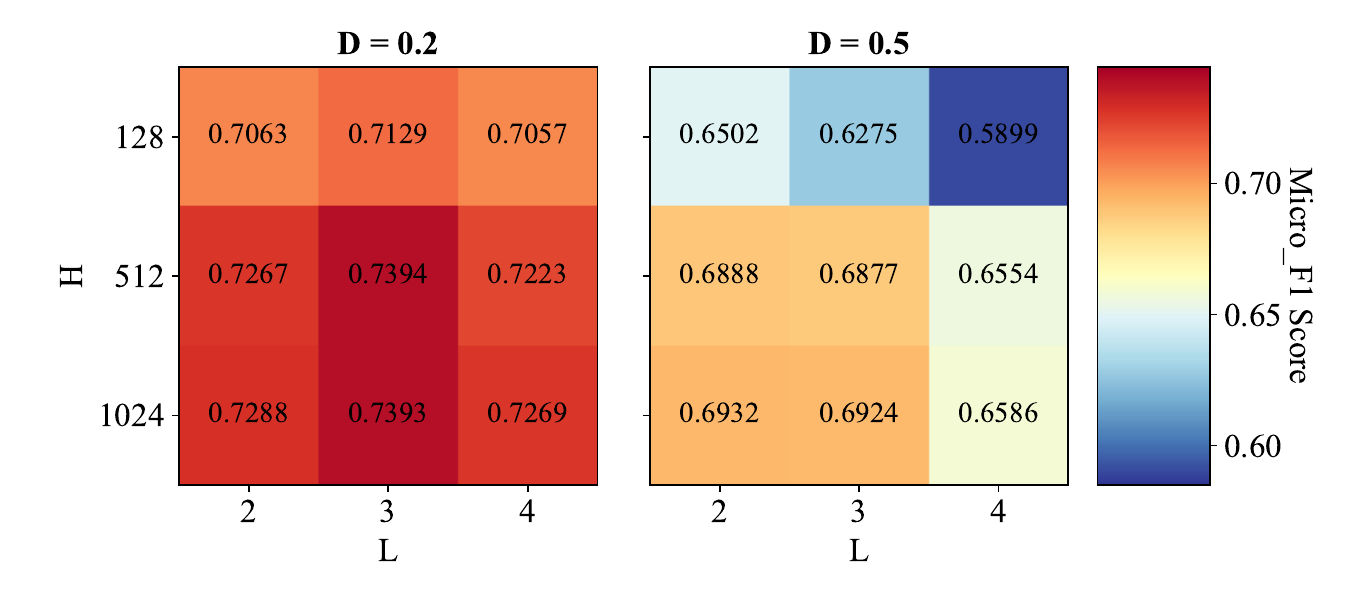}
    }

    \caption{Hyperparameter sensitivity analysis on $L, H, D$.}
    \label{sensitivityfig}
\end{figure*}

\subsection{Ablation studies \textbf{(RQ2)}}
To answer \textbf{RQ2}, we conducted ablation experiments and reported the corresponding coarsening time and node classification performance. We removed each component of the granular-ball graph coarsening in GB-CGNN to study the influence of each component. We maintained the consistency of all other hyperparameters throughout the entire experiment. The experimental results are shown in Table \ref{coarseningtimetable}, Table \ref{acctable}, and Table \ref{F1table}. The ablation research mainly includes \textbf{GB-CGNN-w/o fine-grained binary splitting (-w/o B)} and \textbf{GB-CGNN-w/o coarse-grained initialization (-w/o I)}.

\begin{itemize}
\item \textbf{GB-CGNN-w/o fine-grained binary splitting (-w/o B):} To study the influence of the initialization stage on the construction of multi-granularity granular-ball graph coarsening, we removed the step of binary splitting of granular-balls based on adaptive quality evaluation criteria. 

\item \textbf{GB-CGNN-w/o coarse-grained initialization (-w/o I):} To study the influence of binary splitting of granular-balls based on adaptive quality evaluation criteria on the construction of multi-granularity granular-ball graph coarsening, we removed the initialization stage step.
\end{itemize}

According to the results shown in Table \ref{coarseningtimetable}, Table \ref{acctable}, and Table \ref{F1table}, we can observe that the construction of the granular-ball graph coarsening by these two components contributes to improving the performance of GB-CGNN. The specific analysis is as follows.

\begin{itemize}
\item On the one hand, if there is no fine-grained binary splitting process, the coarsening time of GB-CGNN becomes faster on most datasets. However, on the large-scale graph dataset Amazon, the coarsening time of GB-CGNN has actually slowed down. This situation might occur because when a large graph is directly divided into 1120 subgraphs at one time, the METIS algorithm must perform a complete multi-layer coarsening and refinement on the entire graph. This will lead to each layer's contraction, matching, and merging all being carried out on the large graph, increasing complexity and being cache-unfriendly. On the Amazon dataset, GB-CGNN cannot run due to out of memory. The performance of GB-CGNN will slightly decline on most of the remaining datasets, indicating that detailed granularity binary splitting is of certain importance in the construction of multi-granularity granular-ball coarsening graph. However, on the Citeseer dataset, the performance improved by 5.9\%, which might be due to the unnecessary complexity introduced by this binary splitting process. The complexity of this module could lead to overfitting on the relatively small Citeseer dataset.

\item On the other hand, without a coarse-grained initialization process, the coarsening time of GB-CGNN also becomes faster on all datasets. However, the performance of GB-CGNN shows a significant downward trend on most datasets. Results cannot be produced on the four large-scale graph datasets such as Reddit, Flickr, Yelp, and Amazon. Because at this time, GB-CGNN will not meet the adaptive splitting conditions on almost all datasets and will only generate a few granular-balls, thereby resulting in a significant decline in performance. It is indicated that the initialization of coarse-grained is of great significance in the construction of multi-grained granular-ball coarsening graph. By omitting the coarse-grained initialization, we ignored the effective starting point in the coarsening of the granular-ball graph, which also affected the performance of the model. 

\item Therefore, we can draw the conclusion that the combined and collaborative effect of these two modules enables the model to effectively capture global and local information, thereby enhancing its generalization ability. This indicates that each component plays a crucial role in improving performance.
\end{itemize}

\subsection{Hyper-parameter sensitivity analysis \textbf{(RQ3)}}

\begin{figure}[tb!]
    \centering  
    \includegraphics[width=9.3cm,height=4.3cm]{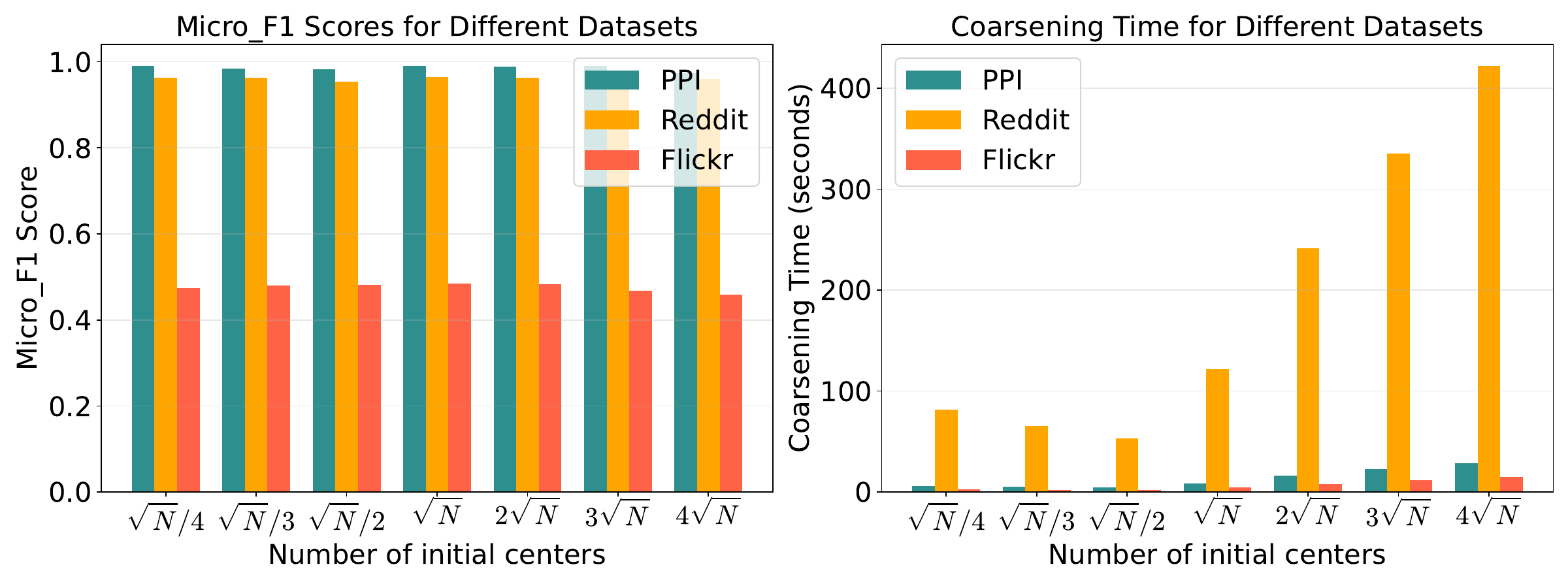}
    \caption{Hyperparameter sensitivity analysis on different initial centers.}
    \label{parameter_N}
\end{figure}

\begin{figure*}[ht]
    \centering
    \subfigure[PubMed]{
        \label{casefig.sub.1}
        \includegraphics[width=0.20\textwidth]{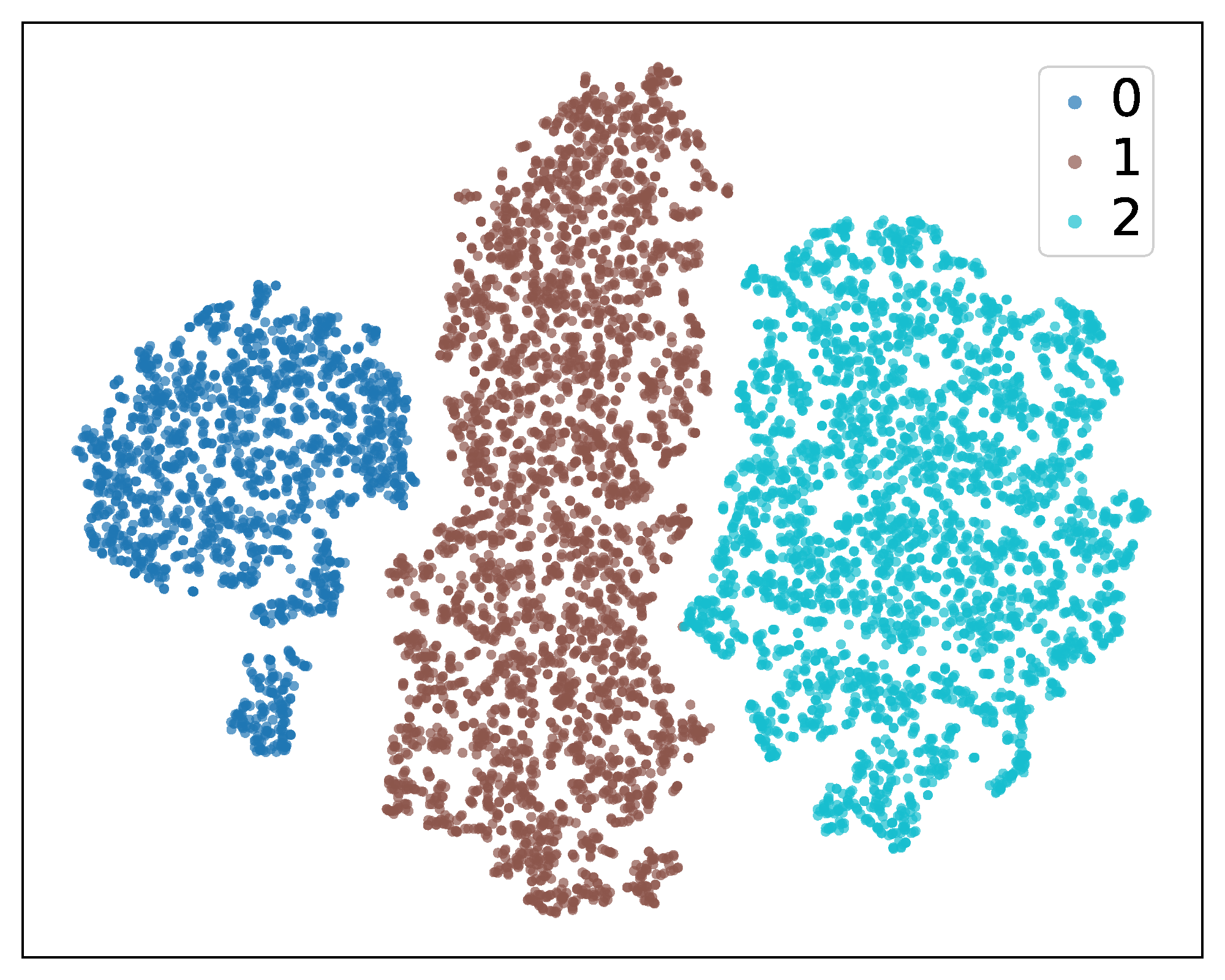}
    }
    \hspace{0.5cm}
    \subfigure[Cocs]{
        \label{casefig.sub.2}
        \includegraphics[width=0.20\textwidth]{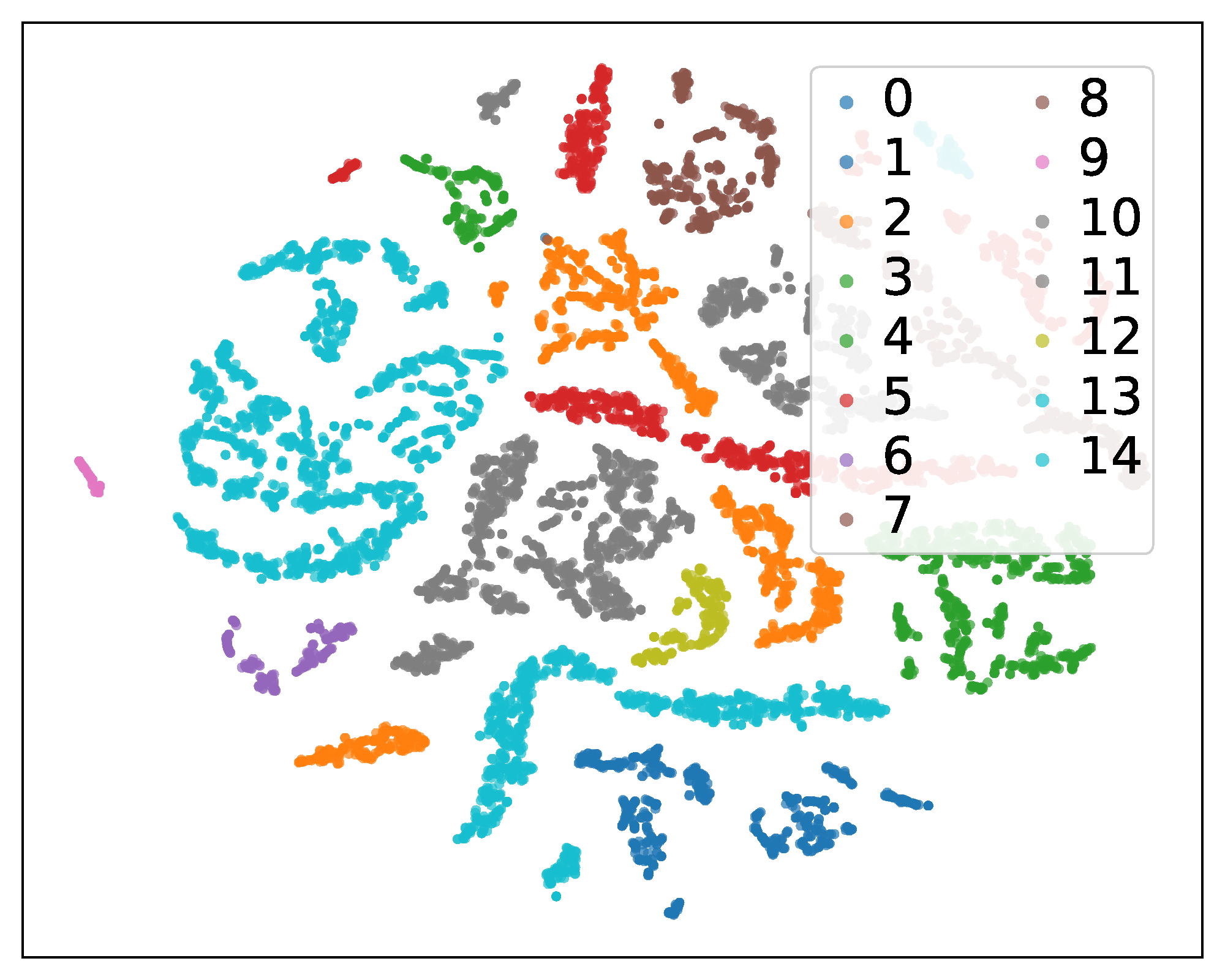}
    }
    \hspace{0.5cm}
    \subfigure[Flickr]{
        \label{casefig.sub.3}
        \includegraphics[width=0.20\textwidth]{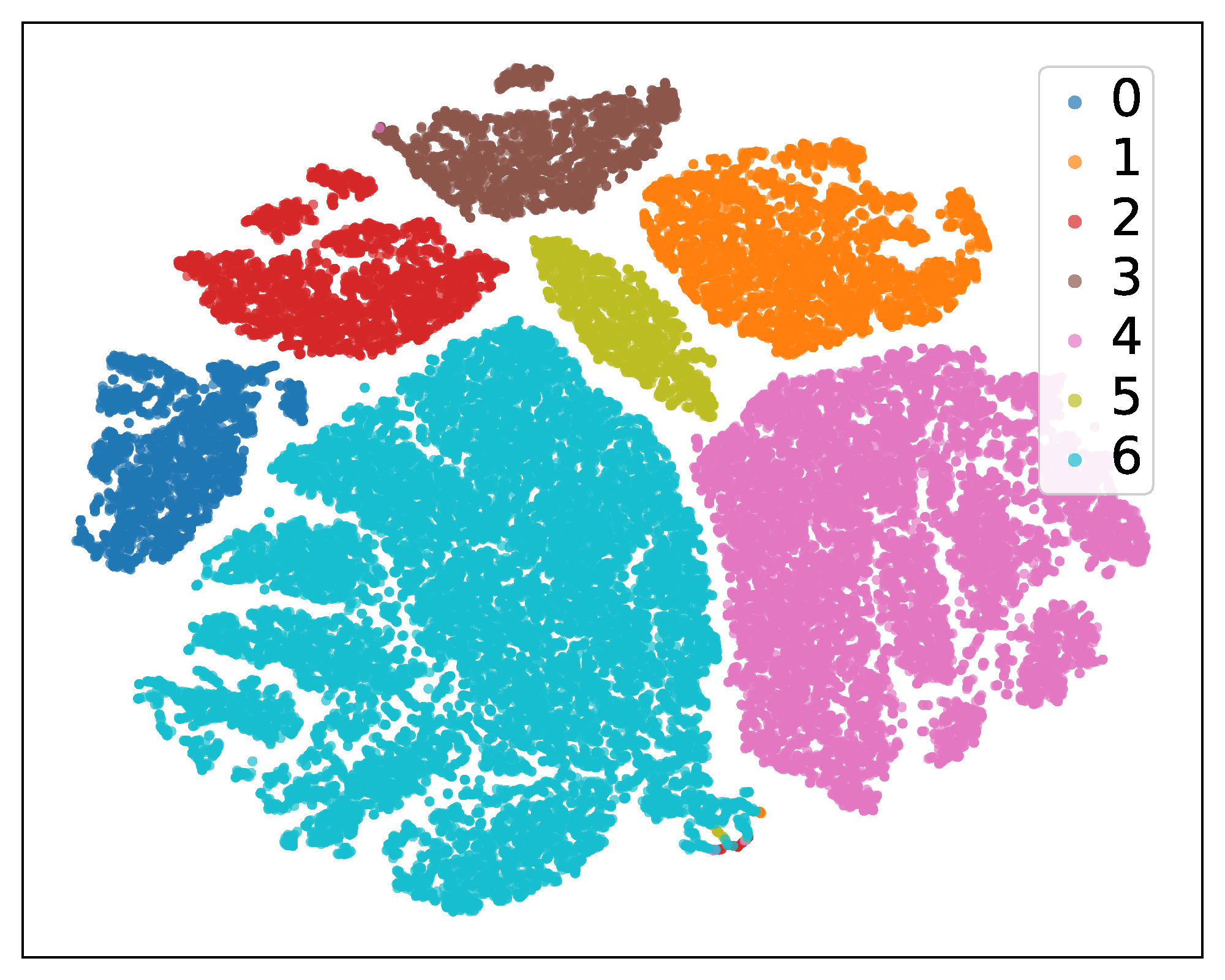}
    }
    \subfigure[Computers]{
        \label{casefig.sub.4}
        \includegraphics[width=0.20\textwidth]{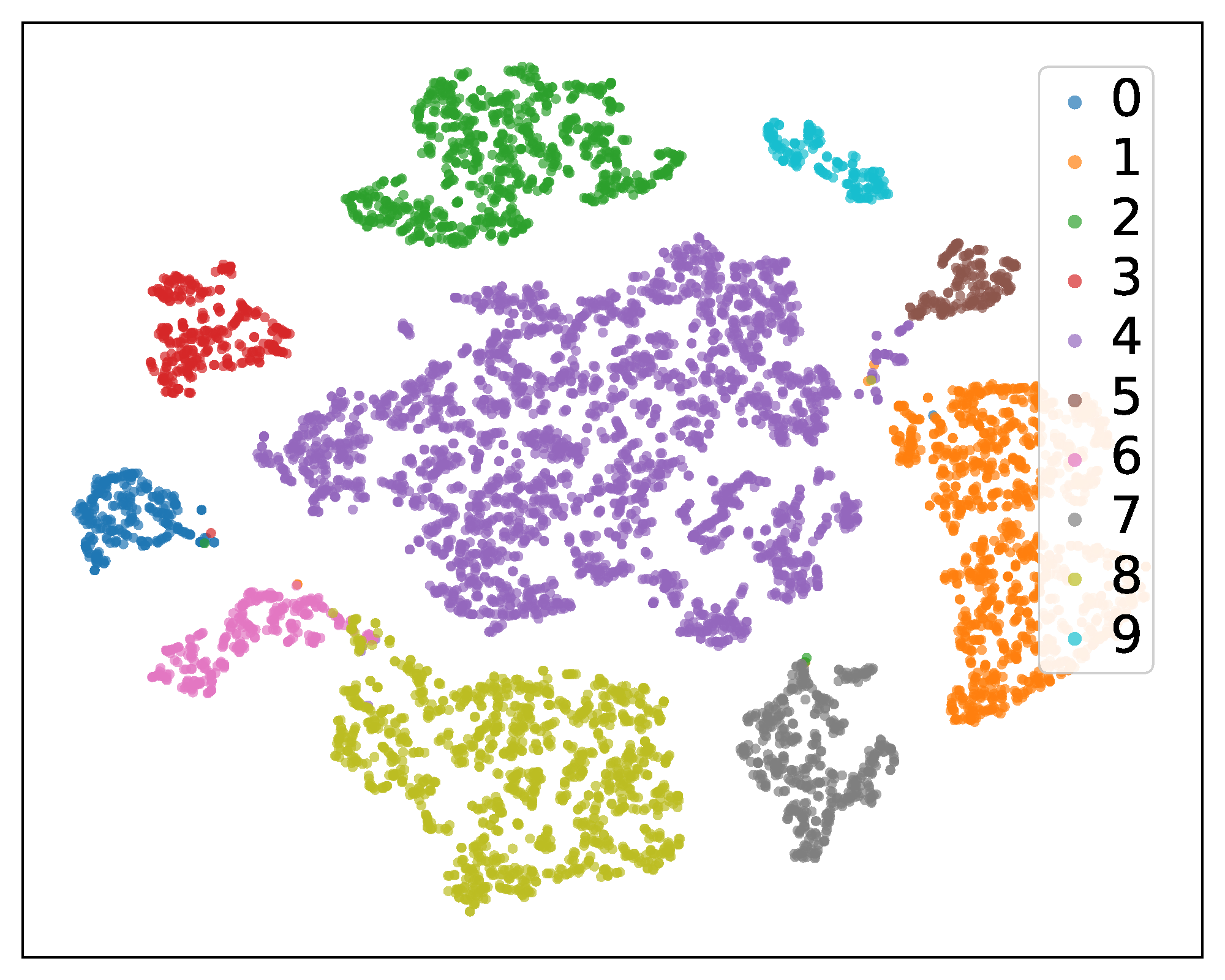}
    }
    \hspace{0.5cm}
    \caption{The node embeddings are visualized on four datasets, with different colors representing distinct node classes.}
    \label{casefig}
\end{figure*}

To answer \textbf{RQ3}, in this subsection, we analyze in detail the key hyperparameters that affect the performance of our proposed GB-CGNN, including $L$, $H$, $D$, and different numbers of initial centers $k$.

\begin{itemize}
\item \textbf{Impact of $L, H, D$.} We explored the influence of $L, H, D$ on the performance of GB-CGNN in the node classification task. Specifically, for the Reddit dataset, we conducted experiments with $L$ values ranging from 3 to 5, $H$ values within the range of \{128, 512, 1024\}, and $D$ values between \{0.2, 0.5\} to observe the changes in the performance of GB-CGNN. For the Yelp dataset, we conducted experiments with $L$ values ranging from 2 to 4, $H$ values within the range of \{512, 1024, 2048\}, and $D$ values between \{0.2, 0.5\} to observe the changes in the performance of GB-CGNN. For the Amazon dataset, we conducted experiments with $L$ values ranging from 2 to 4, $H$ values within the range of \{128, 512, 1024\}, and $D$ values between \{0.2, 0.5\} to observe the changes in the performance of GB-CGNN. For the remaining datasets, we conducted experiments with $L$ values within the range of \{3, 5, 7\}, $H$ values within the range of \{512, 1024, 2048\}, and $D$ values between \{0.2, 0.5\} to determine the changes in the performance of the model. Detailed experimental results are shown in Figure \ref{sensitivityfig}. From the Figure \ref{sensitivityfig}, we can observe that due to the differences in structure, features, etc. among different datasets, parameters $L,H,D$ need to play a synergistic role. This special configuration can match the characteristics of different datasets, achieving an ideal balance between model capacity and computational efficiency, thereby enabling the model to effectively capture complex patterns in the data without overfitting. Ultimately, it enables the model to achieve the best performance. 

\item \textbf{Impact of $k$.} We conducted experiments on different numbers of initial centers ($k$). Figure \ref{parameter_N} shows the $Micro\_F1$ score and coarsening time of GB-CGNN when $k$ of granular-ball was set to $\sqrt{N}/4$, $\sqrt{N}/3$, $\sqrt{N}/2$, $\sqrt{N}$, $2\sqrt{N}$, $3\sqrt{N}$, and $4\sqrt{N}$. As can be seen from the left subgraph of Figure \ref{parameter_N}, the method proposed in this paper still maintains considerable performance even under different number settings, demonstrating a certain degree of stability. Observe the subgraph on the right of Figure \ref{parameter_N}. When the initial number of centers is too large, the corresponding efficiency of coarsening will decrease and the $Micro\_F1$ score will show a downward trend. For instance, on the PPI dataset, the coarsening time for $\sqrt{N}$ centers is 8.69 seconds, while for $4\sqrt{N}$ is 28.71 seconds. The $Micro\_F1$ score has also dropped by approximately 2\%. On the Reddit dataset, although the $Micro\_F1$ score has not decreased significantly, the coarsening time for $\sqrt{N}$ centers is 121.56 seconds, and for $4\sqrt{N}$ centers, it takes 422.35 seconds. Therefore, based on our experiments and empirical evidence from previous work, $\sqrt{N}$ central points were selected as the optimal value, which has been proven to achieve a significantly better balance between accuracy and efficiency.
\end{itemize}

\subsection{Case study \textbf{(RQ4)}}
In this subsection, to answer \textbf{RQ4}, we designed a case study to more intuitively demonstrate the effectiveness of GB-CGNN proposed in this paper. Taking four datasets as examples, including PubMed, Cocs, Flickr and Computers. Figure \ref{casefig} shows the t-SNE visualization of node embedding after training with GB-CGNN. The t-SNE technique can project high-dimensional tensors onto a two-dimensional plane, thereby visually observing the classification effect of GB-CGNN. 

We observed that GB-CGNN can effectively separate nodes of different classes. This is primarily attributed to the algorithm's construction of more compact and locally consistent subgraph regions through an adaptive splitting mechanism from coarse-grained to fine grained, enabling message passing to primarily occur within high-quality local regions, thereby effectively reducing ineffective smoothing and interference from heterogeneous information. Based on this, GB-CGNN can learn node representations that effectively enhance intra-class consistency while simultaneously improving the boundary discriminability between classes, thus facilitating the formation of clear decision boundaries.

\subsection{Impact of different quality evaluations \textbf{(RQ5)}}
To answer \textbf{RQ5}, in this subsection we evaluated the impact of different quality evaluations on the coarsening time and node classification performance of GB-CGNN proposed in this paper. Specifically, it mainly includes the following two experiments: 

\textbf{(I)} Using purity ($P$) as the quality evaluation.

\textbf{(II)} Using $P$ + $AD$ as the quality evaluation. 

The purity $P$ is defined as the ratio of the count of the most frequent labels within the granular-ball to the total number of nodes within the granular-ball, mathematically:

\begin{equation}
P=\frac{max(n_i)}{\widetilde{N}}, i=1,2,\cdots,C,
\end{equation}

\noindent where $n_i$ is the number of nodes with label $i$ in $\mathcal{GB}$, $max(n_i)$ denotes the maximum number of nodes for the same label $i$ in $\mathcal{GB}$, $\widetilde{N}$ is the total number of nodes in $\mathcal{GB}$, and $C$ is the number of classes. Therefore, the higher the $P$, the stronger the consistency of the node labels in the granular-ball $\mathcal{GB}$. In both experiments, the quality threshold was set to 1, meaning that the splitting iteration continued until the quality of each child granular-ball met the predefined threshold $T = 1$. 

The results are shown in Table \ref{rq5}, and we have the following observations.

\begin{itemize}
\item The results in the upper half of Table \ref{rq5} show that when using the first quality evaluation ($P$), the coarsening time on all datasets is slower than that of GB-CGNN. For example, on the Co-phy and Reddit datasets, it is several tens of times slower than GB-CGNN. However, on larger datasets such as Yelp and Amazon, using $P$ as the quality evaluation is not practical, so the coarsening time cannot be calculated due to out of memory. When using the second quality evaluation ($P$ + $AD$), the coarsening time on some datasets is slightly faster than that of GB-CGNN, but for large-scale datasets such as Amazon, the coarsening time is still not as fast as that of GB-CGNN. This indicates that selecting an appropriate quality evaluation during the coarsening process plays an important role in improving the efficiency of model coarsening.

\item The results in the lower half of Table \ref{rq5} show that GB-CGNN exhibits excellent node classification performance. Using the first quality evaluation ($P$) yields poor results, possibly because $P$ tends to result in nodes with the same label into the same granular-ball. However, $P$ ignores the structural relationships between nodes, meaning that splitting granular-balls based on $P$ may lead to the neglect of some important structural information. This is because high consistency in node labels does not necessarily imply a close structural relationship between them in the graph. GCN relies more on graph structural information to propagate node features, so relying solely on $P$ for granular-ball splitting may result in some weakly connected nodes being assigned to the same granular-ball, thereby affecting the final training performance. Therefore, using the second quality evaluation ($P$ + $AD$) takes structural information into account to some extent, resulting in superior classification performance compared to algorithms that only use $P$ as the quality evaluation. GB-CGNN uses $AD$ exclusively for quality evaluation. When using $AD$ as the quality evaluation for granular-ball splitting, it ensures strong connectivity between nodes within each granular-ball, which is particularly important for the GCN model and helps it effectively propagate information during training.
\end{itemize}

In summary, these findings indicate that selecting appropriate quality evaluation is crucial for optimizing the performance of GB-CGNN and capturing multi-granularity structural features of graphs.

\begin{table}[tb!]
    \centering
    \caption{Comparison of coarsening time and performance for different quality evaluations. The results not reported are due to out-of-memory and the best results are bolded. ``I" means using $P$ as the quality evaluation, and ``II" means using $P$ + $AD$ as the quality evaluation. GB: GB-CGNN.}
    \label{rq5} 
    \begin{tabularx}{\linewidth}{@{} *{7}{>{\centering\arraybackslash}X} @{}} 
        \toprule
        \textbf{Evaluation}  & \textbf{Cora} & \textbf{PubMed} & \textbf{Co-phy}& \textbf{Reddit} & \textbf{Yelp} & \textbf{Amazon}\\
        \midrule
        \multicolumn{7}{@{}l}{\textbf{Coarsening Time (unit: second)}} \\ 
        I & 0.59 & 16.77 & 41.24 & 7999.37 & - & - \\
        II & \textbf{0.06} & \textbf{0.41} & \textbf{1.53} & \textbf{116.63} & \textbf{248.20} & 5304.75 \\
        GB & 0.07 & 0.70 & 1.80 & 121.56 & 307.18 & \textbf{4791.81} \\
        \midrule 
        \multicolumn{7}{@{}l}{\textbf{Node Classification Performance}} \\ 
        I & 0.8148 & 0.8676 & 0.9533 & 0.9464 & - & - \\
        II & 0.8556 & 0.8742 & 0.9574 & 0.9567 & 0.5812 & 0.7139 \\
        GB & \textbf{0.8667} & \textbf{0.8813} & \textbf{0.9584} & \textbf{0.9642} & \textbf{0.6268} & \textbf{0.7394} \\
        \bottomrule
    \end{tabularx}
\end{table}

\section{Conclusion}
This paper proposes an Efficient and Scalable Granular-ball Graph Coarsening Method for Large-scale Graph Node Classification (GB-CGNN). GB-CGNN utilizes the multi-granularity structure information of the original graph and simultaneously introduces the METIS algorithm for coarse-grained granular-ball initialization, effectively achieving efficient coarsening of the original graph and solving the limitation of high time complexity of existing coarsening methods. In addition, GB-CGNN accelerates the training efficiency of the GCN algorithm on large-scale graph dataset by training the generated granular-ball subgraphs. Moreover, the experimental results show that the performance of GB-CGNN in the node classification task is superior to that of SOTA algorithms. This demonstrates the method scalability and efficiency, making it a promising solution for large-scale graph processing tasks.

However, GB-CGNN also has its limitations. This method is particularly suitable for graph learning scenarios where multi-granularity structures are prominent, neighborhood information plays a strong discriminative role in downstream tasks, and the training process relies on the construction of high-quality mini-batch subgraphs. For highly irregular graphs or graph tasks that rely more on long-range global dependencies, the advantages of GB-CGNN may be diminished. In the future, further mechanisms will be needed to better preserve global interactions.

\section{Acknowledgment}

We extend our sincere gratitude to everyone whose dedication and efforts contributed to the completion of this work. This research was partially supported by the National Natural Science Foundation of China under Grant Nos. 62222601, 62176033, 62221005, and 61936001; the Key Cooperation Project of the Chongqing Municipal Education Commission under Grant No. HZ2021008; the Scientific and Technological Research Program of the Chongqing Municipal Education Commission under Grant No. KJZD-M202400603; the Project of the Key Laboratory of Tourism Multisource Data Perception and Decision-Making, Ministry of Culture and Tourism, under Grant No. H2023009, as well as by Ant Group, and Chongqing Ant Consumer Finance Co.

\bibliographystyle{IEEEtran}
\bibliography{sample-base}

\vspace{-9 mm}
\begin{IEEEbiography}[{\includegraphics[width=1in,height=1.25in,clip,keepaspectratio]{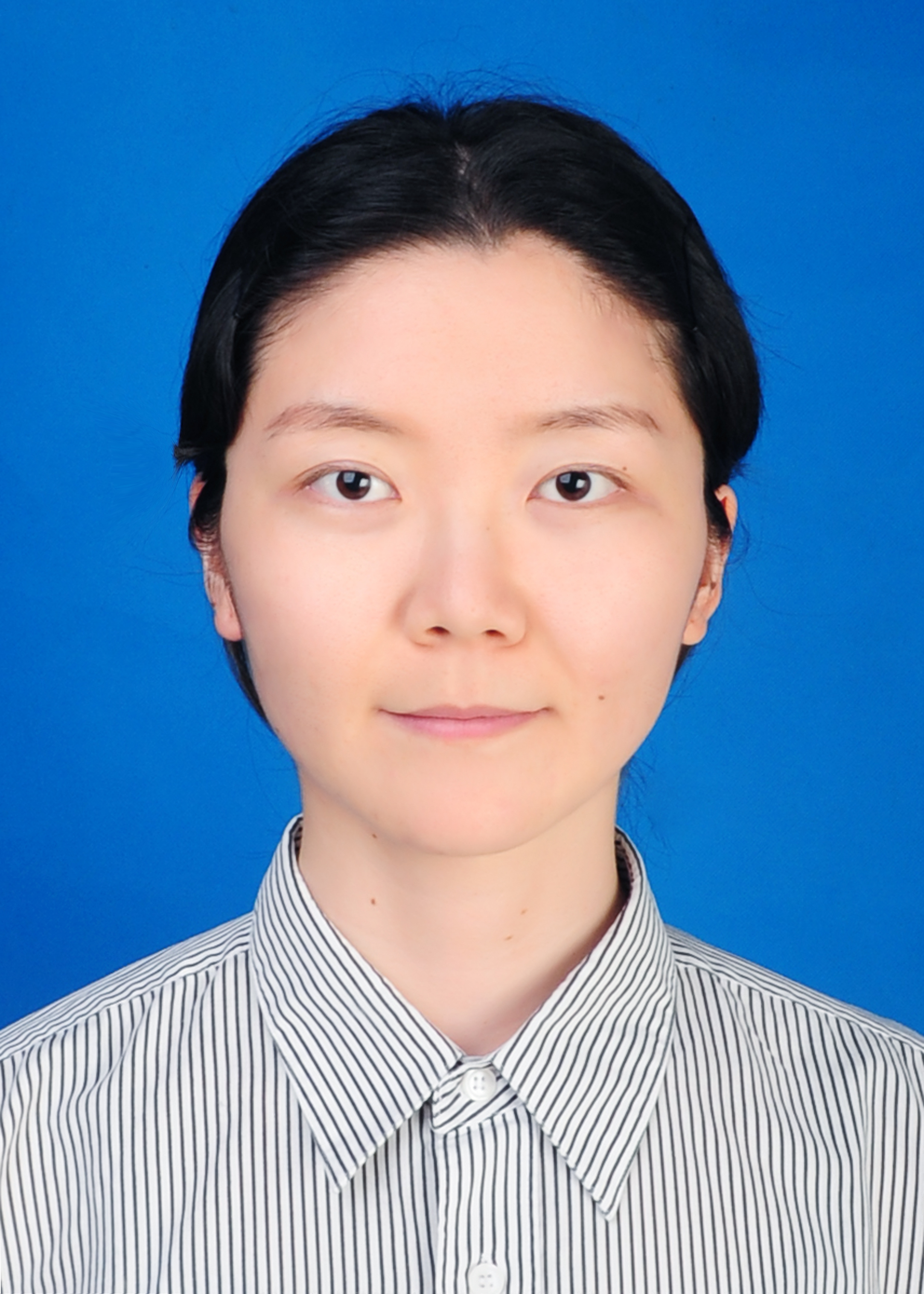}}]{Guan Wang}
received her master's degree from Chongqing University of Posts and Telecommunications in 2024. She is currently a doctoral student in the School of Computer Science and Technology, Chongqing University of Posts and Telecommunications. Her research interest is granular computing and graph representation learning.
\end{IEEEbiography}

\vspace{-9 mm}
\begin{IEEEbiography}[{\includegraphics[width=1in,height=1.25in,clip,keepaspectratio]{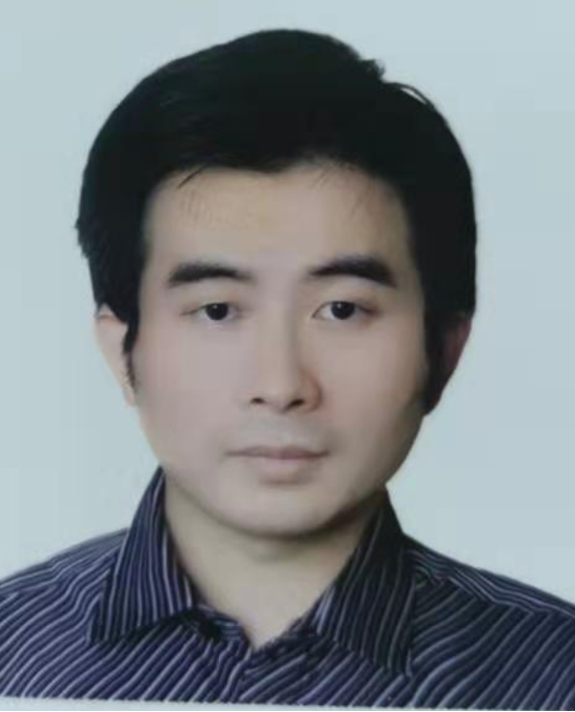}}]{Shuyin Xia$^{*}$}
received his B.S. degree and M.S. degree in Computer science in 2008 and 2012, respectively, both from Chongqing University of Technology in China. He received his Ph.D. degree from College of Computer Science in Chongqing University in China. He is an IEEE Member. Since 2015, he has been working at the Chongqing University of Posts and Telecommunications, Chongqing, China, where he is currently an professor and a Ph.D. supervisor, the executive deputy director of CQUPT - Chongqing Municipal Public Security Bureau - Qihoo 360 Big Data and Network Security Joint Lab. Dr. Xia is the director of Chongqing Artificial Intelligence Association. His research results have expounded at many prestigious journals, such IEEE-TKDE and IS. His research interests include data mining, granular computing, fuzzy rough sets, classifiers and label noise detection.
\end{IEEEbiography}

\vspace{-9 mm}
\begin{IEEEbiography}[{\includegraphics[width=1in,height=1.25in,clip,keepaspectratio]{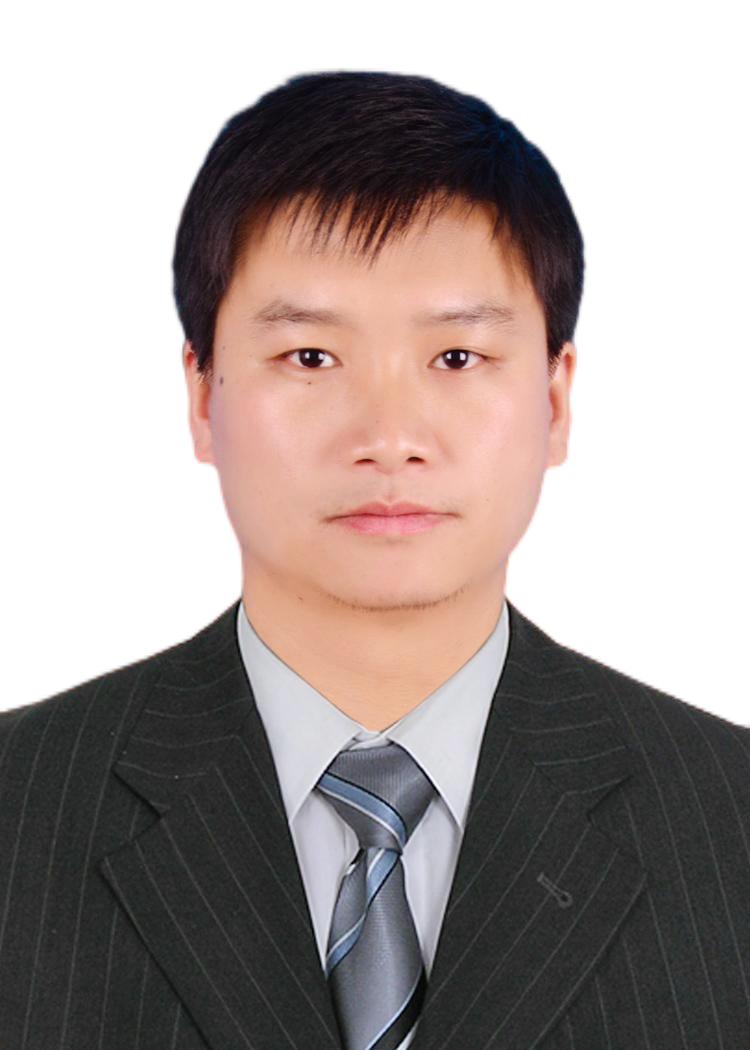}}]{Lei Qian} received his master's degree from Sichuan University in 2017. He is currently a doctoral student in the School of Computer Science and Technology, Chongqing University of Posts and Telecommunications. His research interest is granular computing and graph representation learning.
\end{IEEEbiography}

\vspace{-9 mm}
\begin{IEEEbiography}[{\includegraphics[width=1in,height=1.25in,clip,keepaspectratio]{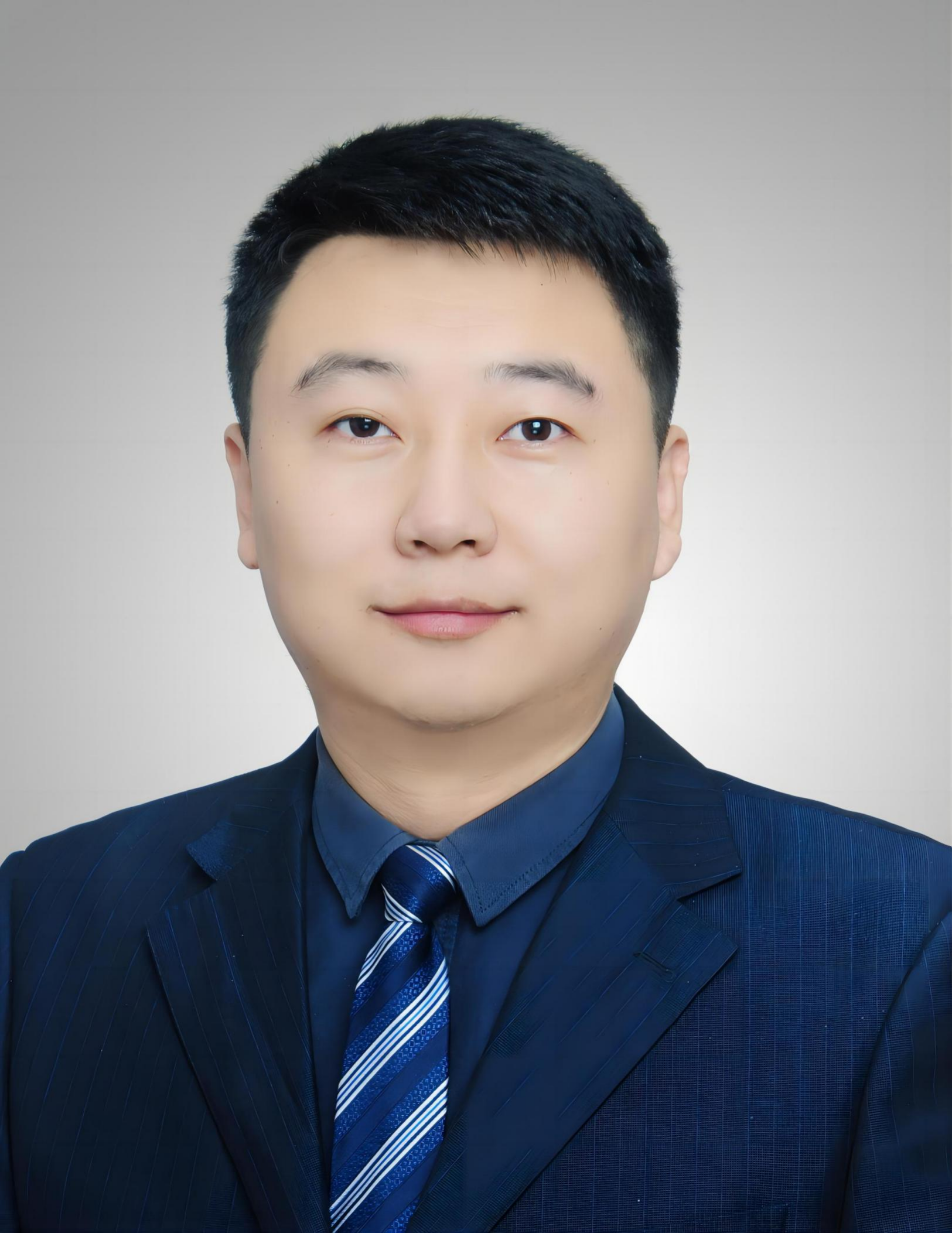}}]{Tao Wu}
received the Ph.D. degree from the University of Electronic Science and Technology of China in 2017. He is currently a Professor and the Head of the Department of Cybersecurity, Chongqing University of Posts and Telecommunications, China. He previously worked at Tencent Technology Co., Ltd., and currently serves as the Executive Deputy Director of the Chongqing Network and Information Security Technology Engineering Laboratory. He has published over 60 papers in high-impact journals and conferences, including IEEE TIE, IEEE TCSVT, IEEE TBD, IEEE TCSS, and PR. His research interests include graph neural networks, graph foundation models, AI security, and adversarial attack and defense. He has served on the Program Committees of Complex Networks 2023, CSE 2024, and CAAI BDSC 2023–2026.
\end{IEEEbiography}

\vspace{-9 mm}
\begin{IEEEbiography}[{\includegraphics[width=1in,height=1.25in,clip,keepaspectratio]{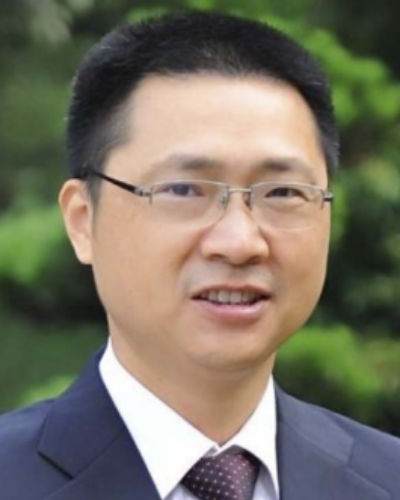}}]{Guoyin Wang(Senior Member, IEEE)}
received the B.S., M.S., and Ph.D. degrees from Xi’an Jiaotong University, Xian, China, in 1992, 1994, and 1996, respectively. He worked at the University of North Texas, and the University of Regina, Canada, as a visiting scholar during 1998-1999. He had worked at the Chongqing University of Posts and Telecommunications during 1996-2024, where he was a professor, the Vice-President of the University, the director of the Chongqing Key Laboratory of Computational Intelligence, the director of the Key Laboratory of Cyberspace Big Data Intelligent Security, Ministry of Education, and the director of the Sichuan-Chongqing Joint Key Laboratory of Digital Economy Intelligence and Security. He was the director of the Institute of Electronic Information Technology, Chongqing Institute of Green and Intelligent Technology, CAS, China, 2011-2017. He has been serving as the President of Chongqing Normal University since June 2024. He is the author of over 20 books, the editor of dozens of proceedings of international and national conferences and has more than 300 reviewed research publications. His research interests include rough sets, granular computing, machine learning, knowledge technology, data mining, neural network, cognitive computing, etc. Dr. Wang was the President of International Rough Set Society (IRSS) 2014-2017, a Vice-President of the Chinese Association for Artificial Intelligence (CAAI) 2014-2025, and a council member of the China Computer Federation (CCF) 2008-2023. He is a supervisor of the Board of Supervisors of CAAI. He is a Fellow of IRSS, CAAI and CCF.
\end{IEEEbiography}

\vspace{-9 mm}
\begin{IEEEbiography}[{\includegraphics[width=1in,height=1.25in,clip,keepaspectratio]{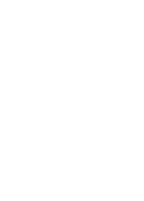}}]{Yi Wang}
received his bachelor's degree in 2010 from 
School of Computer Science, Northwest A\&F University. He is currently employed as a Principal Researcher at Chongqing Ant Consumer Finance Co., Ltd. His research interests including privacy computing, multimodal learning, computer vision.
\end{IEEEbiography}

\vspace{-9 mm}
\begin{IEEEbiography}[{\includegraphics[width=1in,height=1.25in,clip,keepaspectratio]{kb.png}}]{Wei Wang}
received his bachelor's degree in 2012 from 
Huazhong University of Science and Technology.  He is currently employed as a Researcher at Chongqing Ant Consumer Finance Co., Ltd. His research interests including ‌federated learning, deep learning.
\end{IEEEbiography}

\vfill
\end{document}